\documentclass[acmsmall, screen=true, review=false]{acmart}

\AtBeginDocument{%
  \providecommand\BibTeX{{%
    \normalfont B\kern-0.5em{\scshape i\kern-0.25em b}\kern-0.8em\TeX}}}

\usepackage{tikz}
\usetikzlibrary{mindmap,shapes, positioning}
\usepackage{smartdiagram}
\usesmartdiagramlibrary{additions}
\usepackage{forest}
\usetikzlibrary{shadows}
\usepackage{amsmath}
\usepackage{cleveref}
\usepackage{subfigure}
\usetikzlibrary{mindmap}
\usepackage{booktabs}
\usepackage{siunitx}
\usepackage{etoolbox}
\usepackage{nicematrix}
\usepackage{multirow}
\usepackage{multicol}
\usepackage{makecell}
\usepackage{pifont}
\usepackage{enumitem}
\usepackage{pifont}
\usepackage{fontawesome5}
\usepackage{array}
\usepackage{booktabs}
\usepackage{geometry}
\usepackage{graphicx}
\usepackage{ragged2e} 
\usepackage{makecell} 
\usepackage{diagbox} 

\usepackage{booktabs}
\usepackage[T1]{fontenc} 
\usepackage[utf8]{inputenc}
\usepackage[russian, english]{babel}
\usepackage{natbib}
\usepackage{color}
\usepackage{tabularray}
\usepackage{adjustbox}
\usepackage{colortbl}
\usetikzlibrary{positioning}

\definecolor{lightcoral}{rgb}{0.94, 0.5, 0.5}
\definecolor{lightgreen}{rgb}{0.56, 0.93, 0.56}
\definecolor{harvestgold}{rgb}{0.85, 0.57, 0.0}
\definecolor{brightlavender}{rgb}{0.75, 0.58, 0.89}
\definecolor{capri}{rgb}{0.0, 0.75, 1.0}
\definecolor{carminepink}{rgb}{0.92, 0.3, 0.26}
\definecolor{celadon}{rgb}{0.67, 0.88, 0.69}
\definecolor{darkpastelgreen}{rgb}{0.01, 0.75, 0.24}
\definecolor{DeepSkyBlue4}{RGB}{0,104,139}
\definecolor{DeepPurple}{RGB}{48,0,96}
\definecolor{DeepGreen}{RGB}{0,102,0}

\definecolor{softblue}{RGB}{100,149,237}
\definecolor{softgreen}{RGB}{144,238,144}
\definecolor{softpurple}{RGB}{230,190,255}
\definecolor{softorange}{RGB}{255,160,122}
\definecolor{softpink}{RGB}{255,182,193}
\definecolor{majorblue}{RGB}{175,199,232}
\definecolor{majororange}{RGB}{240,145,72}
\definecolor{majoryellow}{RGB}{255,152,150}

\newcommand{\blueref}[1]{\textcolor{blue}{\ref{#1}}}
\usepackage{cellspace}
\begin{document}

\crefformat{section}{\S#2#1#3} 
\crefformat{subsection}{\S#2#1#3}
\crefformat{subsubsection}{\S#2#1#3}
\title{A Survey of Text Watermarking in the Era of Large Language Models}

\author{Aiwei Liu}
\authornote{Both authors contributed equally to this research.}
\email{liuaw20@mails.tsinghua.edu.cn}
\author{Leyi Pan}
\authornotemark[1]
\email{ply20@mails.tsinghua.edu.cn}
\affiliation{
  \institution{Tsinghua University}
  \city{Beijing}
  \country{China}
  }

\author{Yijian Lu}
\email{luyijian@link.cuhk.edu.hk}
\author{Jingjing Li}
\email{lijj@link.cuhk.edu.hk}
\affiliation{
  \institution{The Chinese University of Hong Kong}
  \city{Hong Kong}
  \country{Hong Kong}
  }

\author{Xuming Hu}
\email{xuminghu@hotmail.com}
 \affiliation{
  \institution{The Hong Kong University of Science and Technology (Guangzhou)}
  \city{Guangzhou}
  \country{China}
  }

  \author{Xi Zhang}
\email{zhangx@bupt.edu.cn}
 \affiliation{
  \institution{Beijing University of Posts and Telecommunications}
  \city{Beijing}
  \country{China}
  }
  
\author{Lijie Wen}
\email{wenlj@tsinghua.edu.cn}
\affiliation{
  \institution{Tsinghua University}
  \city{Beijing}
  \country{China}
  }

\author{Irwin King}
\email{king@cse.cuhk.edu.hk}
\affiliation{
  \institution{The Chinese University of Hong Kong}
  \city{Hong Kong}
  \country{Hong Kong}
  }

\author{Hui Xiong}
\email{xionghui@hkust-gz.edu.cn}
\affiliation{
  \institution{The Hong Kong University of Science and Technology (Guangzhou)}
  \city{Guangzhou}
  \country{China}
  }

\author{Philip S. Yu}
\email{psyu@cs.uic.edu}
\affiliation{
  \institution{ University of Illinois Chicago}
  \city{Chicago}
  \country{United States}
  }

\renewcommand{\shortauthors}{Liu, et al.}

\begin{abstract}
Text watermarking algorithms are crucial for protecting the copyright of textual content. Historically, their capabilities and application scenarios were limited. However, recent advancements in large language models (LLMs) have revolutionized these techniques. LLMs not only enhance text watermarking algorithms with their advanced abilities but also create a need for employing these algorithms to protect their own copyrights or prevent potential misuse. This paper conducts a comprehensive survey of the current state of text watermarking technology, covering four main aspects: (1) an overview and comparison of different text watermarking techniques; (2) evaluation methods for text watermarking algorithms, including their detectability, impact on text or LLM quality, robustness under target or untargeted attacks; (3) potential application scenarios for text watermarking technology; (4) current challenges and future directions for text watermarking. This survey aims to provide researchers with a thorough understanding of text watermarking technology in the era of LLM, thereby promoting its further advancement.

\end{abstract}


\keywords{Text Watermark, Large Language Models, Copyright Protection}


\maketitle

\section{Introduction}

\begin{figure} [t]
    \centering
    \subfigure[A description of how LLMs promote the development of text watermarking techniques and broaden their application scenarios. ]{
        \includegraphics[width=0.45\textwidth]{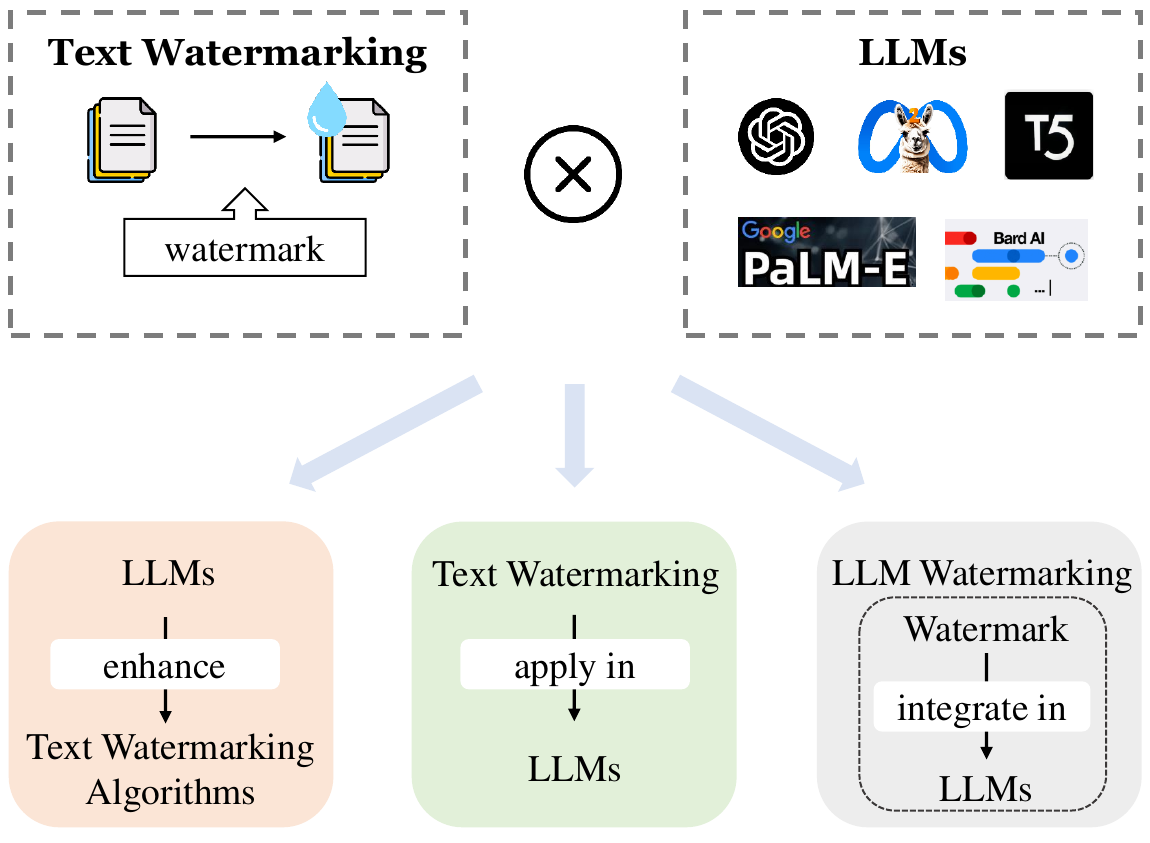} 
        \label{fig:text_watermarking_with_llms}
    }
    \hfill
    \subfigure[Number of publications in the field of text watermarking and LLMs (the data for "Number of Publications in the field of LLMs" is sourced from \citet{zhao2023survey})]{
        \includegraphics[width=0.483\textwidth]{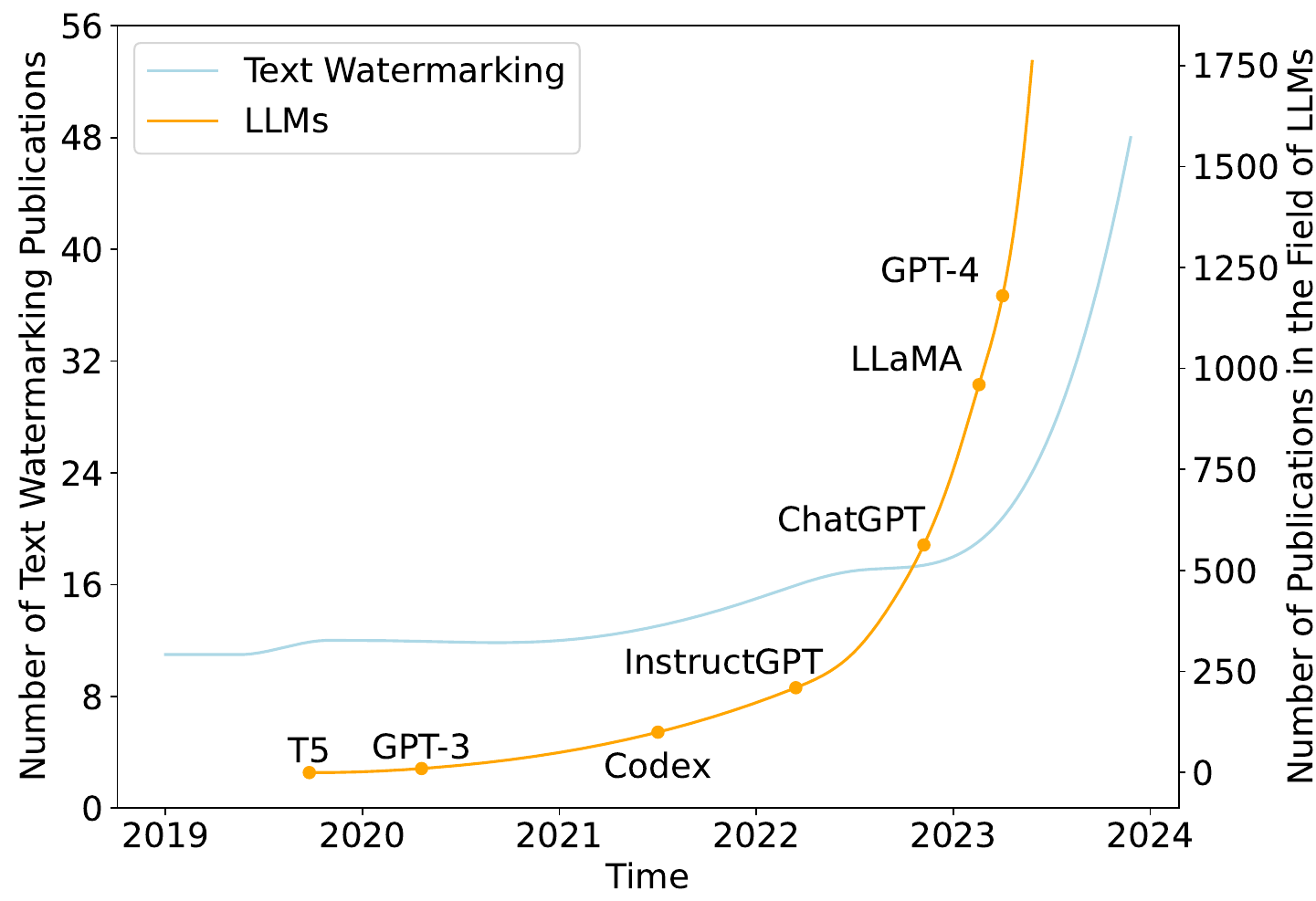} 
        \label{fig:mile_stone}
    }
    \caption{Relationships between the development of text watermarking techniques and LLMs.}
    \vspace{-4mm}
    \label{fig:lead_in}
\end{figure}

Text watermarking involves embedding unique, imperceptible identifiers (watermarks) into textual content. These watermarks are designed to be robust yet inconspicuous, ensuring that the integrity and ownership of the content are preserved without affecting its readability or meaning. Historically, text watermarking has played a crucial role in various domains, from copyright protection and document authentication to preventing plagiarism and unauthorized content distribution \cite{kamaruddin2018review}. With the advancement of Large Language Models (LLMs), both the techniques and application scenarios of text watermarking have seen significant development. As shown in Figure \ref{fig:text_watermarking_with_llms}, this primarily includes the construction of enhanced text watermarking algorithms using LLMs, the application of existing text watermarking algorithms to LLMs, and the exploration of LLM watermarking that directly embeds watermarks during text generation. The flourishing development of LLMs has propelled a thriving research landscape within the realm of text watermarking, as depicted in Figure \color{blue}\ref{fig:mile_stone}\color{black}. Especially with the advent of ChatGPT, text watermarking has notably surged into a research fervor. Specifically, this paper surveys the interplay between LLMs and text watermarking.



\subsection{Why is Text Watermarking Beneficial for LLMs?}
\label{sec:llm-tw}
In recent years, LLMs have made significant progress in the field of natural language processing. As the parameter count of these LLMs continues to increase, their ability to understand and generate language has also substantially improved. Notable examples include GPT \cite{radford2018improving}, BART \cite{lewis2019bart}, T5 \cite{raffel2020exploring}, OPT \cite{zhang2022opt}, LaMDA \cite{thoppilan2022lamda}, LLaMA \cite{touvron2023llama}, and GPT-4 \cite{OpenAI2023GPT4TR}. These LLMs have achieved excellent performance in a variety of downstream tasks, including machine translation \cite{hendy2023good, zhu2020incorporating, costa2022no, hendy2023good}, dialogue systems \cite{hudevcek2023llms, mi2022pangu, thoppilan2022lamda, shuster2022language}, code generation \cite{ni2023lever, vaithilingam2022expectation, nijkamp2022codegen, xu2022systematic}, and other tasks \cite{li2020unsupervised, li2022text, zhang2023benchmarking, thirunavukarasu2023large}. A recent work even suggests that GPT-4 is an early (yet still incomplete) version of an artificial general intelligence (AGI) system \cite{bubeck2023sparks}.
However, the utilization of LLMs introduces several challenges:
\begin{itemize} [labelindent=1pt, labelsep=0.5em, leftmargin=0.6cm]
    \item \textbf{Misuse of LLMs}: LLMs can be exploited by malicious users to create misinformation \cite{chen2023can} or harmful content \cite{perez2022red} and spread on the internet.
    \item \textbf{Intellectual Property Concerns}: Powerful LLMs are vulnerable to model extraction attacks, where attackers extract large amounts of data to train new LLMs \cite{birch2023model}.
\end{itemize}
Adding watermarks to LLM-generated text effectively alleviates these issues. Watermarks enable tracking and detection of LLM-generated text, helping to control potential misuse. Training new LLMs with watermarked text can embed these watermarks, mitigating model extraction attacks.

\subsection{Why are LLMs Beneficial for Text Watermarking?}
\label{sec:tw-llm}
A key challenge in text watermarking is to embed watermarks without distorting the original text's meaning or readability. Traditional methods often fail to modify text without altering its semantics  \cite{atallah2001natural, 10.1145/1178766.1178777, meral2009natural}. The necessity for algorithms to comprehend and control text semantics contributes to this difficulty. However, LLMs significantly alter this landscape. Due to their advanced grasp of language semantics and context, they facilitate sophisticated watermarking approaches that embed watermarks with minimal impact on the text's inherent meaning \cite{abdelnabi2021adversarial, zhang2023remark}. This integration results in more effective and subtle watermarking techniques, preserving the text's original intent while embedding essential watermark features.



\subsection{Why a Survey for Text Watermarking in the Era of LLMs?}
\label{sec:contribution}

Text watermarking technology and LLMs can effectively enhance each other. The interconnection of these two technologies includes the following aspects:
\begin{itemize} [labelindent=0pt, labelsep=0.5em, leftmargin=0.6cm]
    \item \textbf{Watermarking LLM-Generated Text}: Text generated by LLMs can be watermarked using text watermarking algorithms \cite{brassil1995electronic, POR20121075, rizzo2016content, munyer2023deeptextmark, yang2022tracing, yang2023watermarking, yoo2023robust}.
    \item \textbf{Embedding Watermarks via LLMs}: LLMs themselves can be utilized to embed watermarks in texts \cite{abdelnabi2021adversarial, zhang2023remark}.
    \item \textbf{Direct Integration in Text Generation}: Watermark algorithms can be directly incorporated during the text generation process of LLMs \cite{DBLP:conf/icml/KirchenbauerGWK23, zhao2023provable, liu2023semantic, liu2023private, ren2023robust, wu2023dipmark}.
\end{itemize}
However, comprehensive studies exploring text watermarking in the era of LLMs are lacking. Existing surveys predominantly focus on watermarking techniques developed before the advent of LLMs \cite{alkawaz2016concise, kamaruddin2018review}. In this study, we present the first comprehensive survey of text watermarking algorithms in the context of large language models. 
\vspace{3pt}

\textbf{This survey is structured as follows:} Section \blueref{sec:pre} introduces text watermarking definitions and key algorithm properties. Section  \blueref{sec:existing} and Section  \blueref{sec:llm} address two primary text watermarking categories: for existing text and for LLM-generated text. Section  \blueref{sec:evaluation} discusses evaluation metrics for these algorithms, including detectability, quality impact and robustness under watermark attacks. Section  \blueref{sec:application} explores application scenarios, namely copyright protection and AI-generated text detection. Section \blueref{sec:challenges} examines ongoing challenges and potential future research avenues in text watermarking. The survey concludes in Section \blueref{conclusion}.

\section{Preliminaries of Text Watermarking}
\label{sec:pre}

To facilitate the introduction of various text watermarking algorithms as well as its evaluation methods in subsequent sections, this section presents the definition of text watermarking algorithms and outlines the characteristics that an excellent text watermarking algorithm should possess.  The taxonomy of text watermarking algorithms is also introduced in this section.

\begin{figure} [t]
    \centering
    \includegraphics[width=0.98\textwidth]{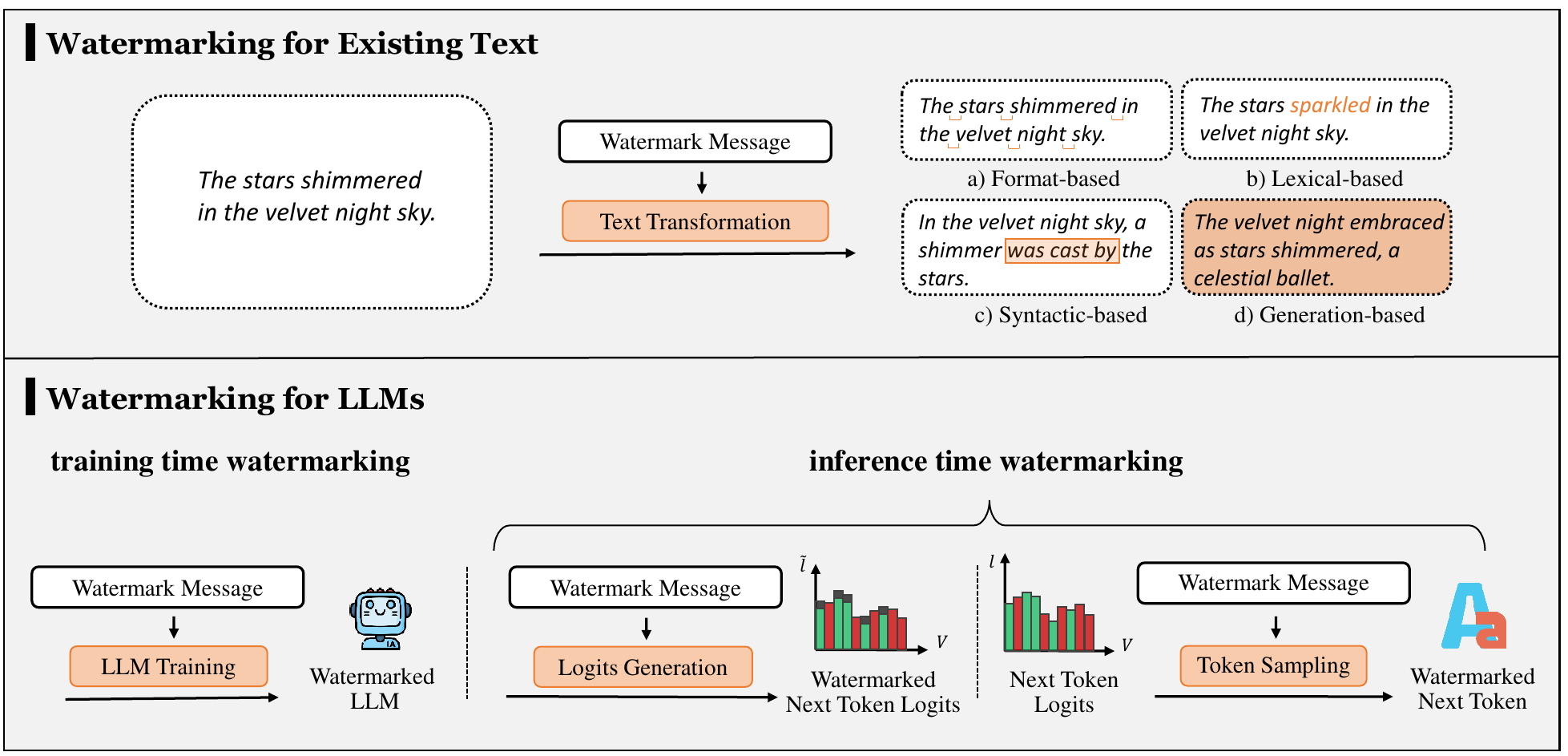}
    \caption{This figure offers a overview of text watermarking techniques. It categorizes watermarking into two main types: for Existing Text and for LLMs. }
    \label{fig:watermarking_method}
    \vspace{-5mm}
\end{figure}

\subsection{Text Watermarking Algorithms}

A text watermarking algorithm typically comprises two components: a watermark generator $\mathcal{A}$, and a watermark detector $\mathcal{D}$. The watermark generator $\mathcal{A}$ takes a text $\mathbf{x}$ and a watermark message $w$ as inputs and outputs a watermarked text $\mathbf{t}$, expressed as $\mathcal{A}(\mathbf{x}, w) = \mathbf{t}$.

This watermarked text $\mathbf{t}$ is either in a different form but semantically equivalent to the original text $\mathbf{x}$ (\cref{sec:existing}) or a newly generated text in response to $\mathbf{x}$ (\cref{sec:llm}), particularly in contexts like prompts for LLMs.
The watermark message, denoted as $w$, can be a zero-bit watermark, signifying merely its presence or absence, or a multi-bit watermark, embedding detailed, customized information. The phrase "watermark payload"  denote the information volume conveyed by $w$.

For the watermark detector $\mathcal{D}$, its input is any text $\mathbf{t}$, and its output is its predicted watermark message for the text, denoted as $\mathcal{D}(\mathbf{t}) = w$. If the output is None, it implies that the text contains no watermark information.

\subsection{Connection with Related Concepts}

To further clarify the scope of text watermarking discussed in this work, this section distinguishes the mentioned text watermarking from other related concepts.
\begin{itemize}[labelindent=0pt, labelsep=0.5em, leftmargin=*]
    \item \textbf{Steganography}: Both steganography \cite{csahin2021review} and text watermarking are important methods of information hiding. While similar, steganography typically requires higher capacity for hidden information, whereas watermarking prioritizes robustness to further text modifications.
    \item \textbf{LLM Watermarking}: The concept of LLM watermarking includes all forms of watermarking added to LLMs, such as their parameters \cite{uchida2017embedding}, output embeddings \cite{peng2023you}, and text \cite{DBLP:conf/icml/KirchenbauerGWK23}. This work focuses solely on watermarking applied to the output text of LLMs.
\end{itemize}

\subsection{Key Characteristics of Text Watermarking Algorithms}

To further enhance understanding of the text watermarking concept, this section introduces two key characteristics: low impact on text quality and robustness to watermark removal attacks.



 \textbf{Low Impact on Text Quality.} The quality of text should not substantially decrease after adding a watermark. Let $\mathcal{A}(\mathbf{x}, \emptyset)$ represent the text generated without a watermark. When $\mathbf{x}$ is the target text (\cref{sec:existing}), the output remains $\mathbf{x}$. For a prompt given to a LLM (\cref{sec:llm}), it denotes the LLM's output without a watermark. An effective watermarking algorithm ensures minimal impact on text quality:
\begin{equation}
    \forall w_i, \, \mathcal{R}(\mathcal{A}(\mathbf{x}, \emptyset), \mathcal{A}(\mathbf{x}, w_i)) < \delta,
    \label{eq:quality}
\end{equation}
 where $\mathcal{R}$ is a function evaluating text quality from multiple perspectives, as will be discussed in Section \blueref{sec:evaluation}. $\delta$ represents a threshold. If the difference in the evaluated scores of two texts is less than this threshold, they are considered to be of similar quality.

  \textbf{Robustness to watermark removal attack.} For a text watermarking algorithm, it is crucial that the watermarked text can still be detected after some modifications. We use an operation $\mathcal{U}$ to denote the watermark removal operations, which will be detailed in Section \blueref{sec:evaluation}. If a watermarking algorithm is robust against watermark removal attacks, it should satisfy the following conditions:
\begin{equation}
     \forall w_i, \forall \mathbf{t} = \mathcal{A}(\mathbf{x}, w_i), \, P(\mathcal{D}(\mathcal{U}(\mathbf{t})) = w_i) > \beta,
\end{equation}
 where $\beta$ is a threshold. If the probability of correctly detecting a watermarked text after text modification exceeds $\beta$, the algorithm is deemed sufficiently robust.

Additionally, there are other important characteristics of text watermarking algorithms which will be discussed in detail in Section \ref{sec:evaluation}.

\subsection{Taxonomy of Text Watermarking Algorithms}

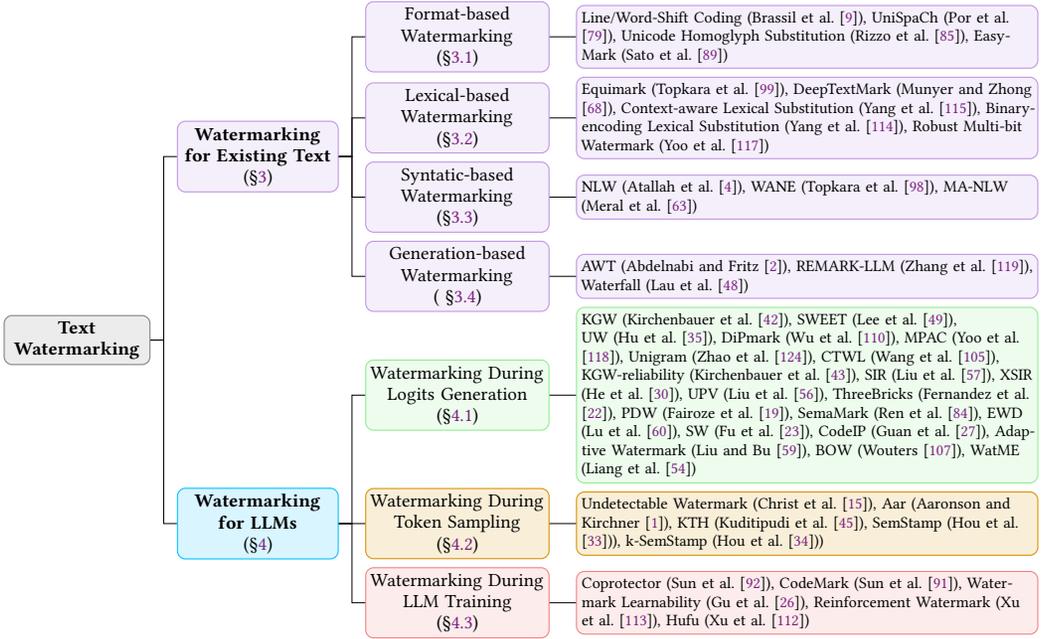
\begin{figure}[t]
\centering
\tikzset{
        my node/.style={
            draw,
            align=center,
            thin,
            text width=1.2cm, 
            rounded corners=3,
        },
        my leaf/.style={
            draw,
            align=left,
            thin,
            text width=8.5cm, 
            rounded corners=3,
        }
}
\forestset{
  every leaf node/.style={
    if n children=0{#1}{}
  },
  every tree node/.style={
    if n children=0{minimum width=1em}{#1}
  },
}
\begin{forest}
    nonleaf/.style={font=\scriptsize},
     for tree={%
        every leaf node={my leaf, font=\tiny},
        every tree node={my node, font=\tiny, l sep-=4.5pt, l-=1.pt},
        anchor=west,
        inner sep=2pt,
        l sep=10pt, 
        s sep=3pt, 
        fit=tight,
        grow'=east,
        edge={ultra thin},
        parent anchor=east,
        child anchor=west,
        if n children=0{}{nonleaf}, 
        edge path={
            \noexpand\path [draw, \forestoption{edge}] (!u.parent anchor) -- +(5pt,0) |- (.child anchor)\forestoption{edge label};
        },
        if={isodd(n_children())}{
            for children={
                if={equal(n,(n_children("!u")+1)/2)}{calign with current}{}
            }
        }{}
    }
    [\textbf{Text \\ Watermarking}, draw=gray, fill=gray!15, text width=1.8cm, text=black
    [\textbf{Watermarking for Existing Text }\\ ({\cref{sec:existing}}), color=brightlavender, fill=brightlavender!15, text width=2cm, text=black
            [Format-based Watermarking \\ (\cref{sec:format}), color=brightlavender, fill=brightlavender!15, text width=2.3cm, text=black
                [Line/Word-Shift Coding ({\citet{brassil1995electronic}),  UniSpaCh (\citet{POR20121075}), Unicode Homoglyph Substitution (\citet{rizzo2016content}), EasyMark (\citet{sato2023embarrassingly})}, color=brightlavender, fill=brightlavender!15, text width=6.0cm, text=black]
            ]
            [Lexical-based Watermarking \\ (\cref{sec:lexical}), color=brightlavender, fill=brightlavender!15, text width=2.3cm, text=black
                [ Equimark ({\citet{topkara2006hiding}), DeepTextMark (\citet{munyer2023deeptextmark}), Context-aware Lexical Substitution (\citet{yang2022tracing}), 
                         Binary-encoding Lexical Substitution (\citet{yang2023watermarking}), Robust Multi-bit Watermark (\citet{yoo2023robust})
                    }, color=brightlavender, fill=brightlavender!15, text width=6.0cm, text=black]
            ],
            [Syntatic-based Watermarking \\ (\cref{sec:syntactic}), color=brightlavender, fill=brightlavender!15, text width=2.3cm, text=black
                [ {NLW (\citet{atallah2001natural}),  WANE (\citet{10.1145/1178766.1178777}), MA-NLW (\citet{meral2009natural})
                    }, color=brightlavender, fill=brightlavender!15, text width=6.0cm, text=black]
            ],
            [Generation-based Watermarking  \\ ( \cref{sec:generation}), color=brightlavender, fill=brightlavender!15, text width=2.3cm, text=black
                [ { AWT (\citet{abdelnabi2021adversarial}), REMARK-LLM (\citet{zhang2023remark}), Waterfall (\citet{lau2024waterfall})
                    }, color=brightlavender, fill=brightlavender!15, text width=6.0cm, text=black]
            ]
        ]
        [\textbf{Watermarking for LLMs} \\ (\cref{sec:llm}), color=capri, fill=capri!15, text width=2cm, text=black
            [Watermarking During Logits Generation  \\ (\cref{sec:logits-watermark}), color=lightgreen, fill=lightgreen!15, text width=2.3cm, text=black
                [{KGW (\citet{DBLP:conf/icml/KirchenbauerGWK23}), SWEET (\citet{lee2023wrote}), \\ UW (\citet{hu2023unbiased}), DiPmark (\citet{wu2023dipmark}), MPAC (\citet{yoo2023advancing}), Unigram (\citet{zhao2023provable}), CTWL (\citet{wang2023towards}), KGW-reliability (\citet{kirchenbauer2023reliability}), SIR (\citet{liu2023semantic}), XSIR (\citet{he2024can}), UPV (\citet{liu2023private}), ThreeBricks (\citet{fernandez2023three}),  
                 PDW (\citet{cryptoeprint:2023/1661}), SemaMark (\citet{ren2023robust}), EWD (\citet{lu2024entropy}), SW (\citet{fu2024watermarking}), CodeIP (\citet{guan2024codeip}), Adaptive Watermark (\citet{liu2024adaptive}), BOW (\citet{wouters2023optimizing}), WatME (\citet{liang2024watme})}, color=lightgreen, fill=lightgreen!15, text width=6.0cm, text=black]
            ],
            [Watermarking During Token Sampling \\ (\cref{sec:token-watermark}), color=harvestgold, fill=harvestgold!15, text width=2.3cm, text=black
                [{Undetectable Watermark (\citet{christ2024undetectable}), Aar (\citet{aronsonpowerpoint}), KTH (\citet{kuditipudi2023robust}), SemStamp (\citet{hou2023semstamp})), k-SemStamp (\citet{hou2024k}))}, color=harvestgold, fill=harvestgold!15, text width=6.0cm, text=black]
            ],
            [{Watermarking During \\ LLM Training} \\ (\cref{sec:train-watermark}), color=lightcoral, fill=lightcoral!15, text width=2.3cm, text=black
                [{Coprotector (\citet{sun2022coprotector}), CodeMark (\citet{sun2023codemark}), Watermark Learnability (\citet{gu2024on}),
                Reinforcement Watermark
                (\citet{xu2024learning}),
                Hufu
                (\citet{xu2024hufu})  }, color=lightcoral, fill=lightcoral!15, text width=6.0cm, text=black]
            ],
        ]
    ]
    \end{forest}
\caption{Taxonomy of text watermarking methods.}
\label{fig_taxonomy_of_methods}
\vspace{-3mm}
\end{figure}

To facilitate the organization of different text watermarking algorithms in Section \color{blue}\ref{sec:existing} \color{black} and Section \color{blue}\ref{sec:llm}\color{black}, this section provides an overview of our summarized taxonomy of text watermarking algorithms.
Figure \blueref{fig:watermarking_method} categorizes text watermarking methods into two primary types. The first type, \textbf{Watermarking for Existing Text}, embeds watermarks by post-processing pre-existing texts, as elaborated in Section \blueref{sec:existing}. This technique typically utilizes semantically invariant transformations for watermark integration. The second type, \textbf{Watermarking for Large Language Models}, involves modifying LLMs, further detailed in Section \blueref{sec:llm}. This method either embeds specific features during LLM training, or alters the inference process, producing watermarked text from the input prompt. Figure \blueref{fig_taxonomy_of_methods} presents a more detailed taxonomy of all text watermarking methods.

\section{Watermarking for Existing Text}
\label{sec:existing}

Watermarking for existing text involves modifying a generated text to produce a watermarked text. Based on the granularity of modifications, these methods are primarily categorized into four types:
format-based watermarking (\cref{sec:format}), lexical-based watermarking (\cref{sec:lexical}), syntactic-based watermarking (\cref{sec:syntactic}) and generation-based watermarking (\cref{sec:generation}). 

\subsection{Format-based Watermarking}
\label{sec:format}

Format-based watermarking, inspired by image watermarking \cite{begum2020digital}, changes the text format rather than its content to embed watermarks. For example, \citet{brassil1995electronic} introduced line-shift and word-shift coding by adjusting text lines and words vertically and horizontally. The detection process identifies shifts by measuring distances between text line profiles or word column profiles. However, this method is limited to image-formatted text and does not embed a watermark in the text string.

To address this, Unicode codepoint insertion/replacement methods have emerged. \citet{POR20121075} developed \texttt{UniSpach}, which inserts Unicode space characters in various text spacings. \citet{rizzo2016content} introduced a unicode homoglyph substitution method, using visually similar but differently coded text symbols (e.g., U+0043 and U+216d for 'C', U+004c and U+216c for 'L'). Recently, \texttt{EasyMark} \citep{sato2023embarrassingly}, a family of simple watermarks, was proposed. It includes \texttt{WhiteMark}, which replaces a whitespace (U+0020) with another whitespace codepoint (e.g., U+2004); \texttt{VariantMark}, which uses Unicode variation selectors for CJK texts; and \texttt{PrintMark}, which embeds watermark messages in printed texts using ligatures or slightly different whitespace lengths. The detection process involves searching for specific inserted codepoints.

Though format-based watermarking methods can embed substantial payloads without changing textual content, their format modifications can be noticeable. \citet{POR20121075} noted the \texttt{DASH} attack's ability to highlight these changes. Thus, these methods are vulnerable to removal through canonicalization \cite{boucher2022bad}, such as resetting line spacing and replacing specific codepoints. Additionally, these detectable formats may be exploited for watermark forgery, reducing detection efficacy.

\subsection{Lexical-based Watermarking}  \label{sec:lexical} 

Format-based watermarking approaches, which only modify text's superficial format, are prone to be spotted, making them easily removable through reformatting. This highlights the need for investigating deeper watermark embedding methods in text. A number of studies have adopted word-level modifications, where selected words are replaced with their synonyms without altering the sentence's syntactic structure \citep{topkara2006hiding, munyer2023deeptextmark, yang2022tracing, yang2023watermarking, yoo2023robust}. These are known as lexical-based watermarking approaches. \citet{topkara2006hiding} introduced a synonym substitution method, employing \texttt{WordNet} \citep{fellbaum1998wordnet} as the synonym source. The watermark detection replicates the embedding process, applying inverse rules for message extraction. \citet{munyer2023deeptextmark} enhanced semantic modeling by using a pretrained \texttt{Word2Vec} model, converting selected words into vectors and identifying n-nearest vectors as replacement candidates. They employed a binary classifier with a pretrained \texttt{BERT} model and transformer blocks for watermark detection.

The aforementioned watermarking methods relying on context-independent synonym substitution (\texttt{WordNet} \& \texttt{Word2Vec}) often neglect the context of target words, potentially compromising sentence semantics and text quality. To address this, context-aware lexical substitution has been incorporated into text watermarking.
\citet{yang2022tracing} introduced a BERT-based infill model for generating contextually appropriate lexical substitutions. The watermark detection algorithm parallels the generation process, identifying watermark-bearing words, generating substitutes, and applying inverse rules for message extraction.
\citet{yang2023watermarking} streamlined watermark detection by encoding each word with a random binary value and substituting bit-0 words with context-based synonyms representing bit-1. As non-watermarked text adheres to a Bernoulli distribution, altered during watermarking, statistical tests can effectively detect watermarks.
\citet{yoo2023robust} enhanced robustness against removal attacks by fine-tuning a BERT-based infill model with keyword-preserving and syntactically invariant corruptions, achieving superior robustness compared to earlier methods.

\begin{figure}[t]
\centering
\tikzset{
        my node/.style={
            draw,
            align=center,
            thin,
            text width=1.2cm, 
            rounded corners=3,
        },
        my leaf/.style={
            draw,
            align=left,
            thin,
            text width=8.5cm, 
            rounded corners=3,
        }
}
\forestset{
  every leaf node/.style={
    if n children=0{#1}{}
  },
  every tree node/.style={
    if n children=0{minimum width=1em}{#1}
  },
}
\begin{forest}
    nonleaf/.style={font=\scriptsize},
     for tree={%
        every leaf node={my leaf, font=\tiny},
        every tree node={my node, font=\tiny, l sep-=4.5pt, l-=1.pt},
        anchor=west,
        inner sep=2pt,
        l sep=10pt, 
        s sep=3pt, 
        fit=tight,
        grow'=east,
        edge={ultra thin},
        parent anchor=east,
        child anchor=west,
        if n children=0{}{nonleaf}, 
        edge path={
            \noexpand\path [draw, \forestoption{edge}] (!u.parent anchor) -- +(5pt,0) |- (.child anchor)\forestoption{edge label};
        },
        if={isodd(n_children())}{
            for children={
                if={equal(n,(n_children("!u")+1)/2)}{calign with current}{}
            }
        }{}
    }
    [\textbf{Watermarking for Existing Text}, draw=gray, fill=gray!15, text width=2.7cm, text=black
    [\textbf{Format-based \\ Watermarking (\cref{sec:format})}, color=brightlavender, fill=brightlavender!15, text width=3cm, text=black
                [Text Layout Modification, color=brightlavender, fill=brightlavender!15, text width=3cm, text=black            [\citep{brassil1995electronic}, color=brightlavender, fill=brightlavender!15, text width=1.0cm, text=black]
                ]
                [Unicode-based Substitution,
                color=brightlavender, fill=brightlavender!15, text width=3cm, text=black
                [\citep{POR20121075, rizzo2016content, sato2023embarrassingly}, color=brightlavender, fill=brightlavender!15, text width=1.0cm, text=black]
                ]
        ]
        [\textbf{Lexical-based \\ Watermarking (\cref{sec:lexical})}, color=brightlavender, fill=brightlavender!15, text width=3cm, text=black
            [Context-independent \\ Lexical Substitution, color=brightlavender, fill=brightlavender!15, text width=3cm, text=black
                [\citep{topkara2006hiding,  munyer2023deeptextmark}, color=brightlavender, fill=brightlavender!15, text width=1.0cm, text=black]
                ]
                [Context-aware \\ Lexical Substitution,
                color=brightlavender, fill=brightlavender!15, text width=3cm, text=black
                [\citep{yang2022tracing, yang2023watermarking, yoo2023robust},  color=brightlavender, fill=brightlavender!15, text width=1.0cm, text=black]
                ]
        ]
        [\textbf{Syntactic-based \\ Watermarking (\cref{sec:syntactic})}, color=brightlavender, fill=brightlavender!15, text width=3cm, text=black
                [Sentence Structure Transformations, color=brightlavender, fill=brightlavender!15, text width=3cm, text=black
                [\citep{atallah2001natural, 10.1145/1178766.1178777}, color=brightlavender, fill=brightlavender!15, text width=1.0cm, text=black]
                ]
                [Language-specific \\ Morphosyntactic Tools,
                color=brightlavender, fill=brightlavender!15, text width=3cm, text=black
                [\citep{meral2009natural}, color=brightlavender, fill=brightlavender!15, text width=1.0cm, text=black]
                ]
        ]
        [\textbf{Generation-based \\ Watermarking (\cref{sec:generation})}, color=brightlavender, fill=brightlavender!15, text width=3cm, text=black
                [Utilizing Self-designed Neural Networks, color=brightlavender, fill=brightlavender!15, text width=3cm, text=black
                [\citep{abdelnabi2021adversarial,zhang2023remark}, color=brightlavender, fill=brightlavender!15, text width=1.0cm, text=black]
                ]
                [Utilizing Watermarked LLMs,
                color=brightlavender, fill=brightlavender!15, text width=3cm, text=black
                [\citep{lau2024waterfall}, color=brightlavender, fill=brightlavender!15, text width=1.0cm, text=black]
                ]
        ]
        ],
    ]
]
\end{forest}
\caption{Taxonomy of watermarking for existing text.}
\label{fig:taxonomy_of_existing}
\vspace{-3mm}
\end{figure}
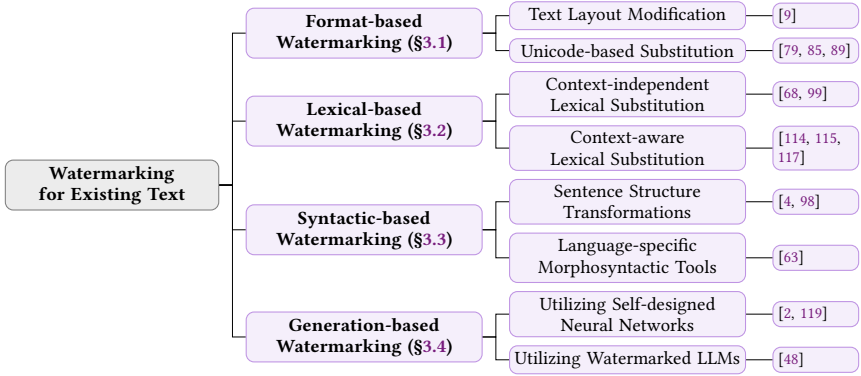

\subsection{Syntactic-based Watermarking}
\label{sec:syntactic}
The lexical-based methods aim to embed watermarks by substituting specific words, maintaining the sentence's syntax. Yet, these approaches, relying exclusively on lexical substitution, might not be robust against straightforward watermark removal tactics like random synonym replacement. Consequently, several studies have explored embedding watermarks in a manner that resists removal, notably by modifying the text's syntax structure. These methods are known as syntactic-based watermarking approaches. 
\citet{atallah2001natural} introduced three typical syntax transformations\textendash \textit{Adjunct Movement}, \textit{Clefting} and \textit{Passivization}\textendash to embed watermark messages.


Each transformation type is assigned a unique message bit: \textit{Adjunct Movement} to 0, \textit{Clefting} to 1, and \textit{Passivization} to 2. In watermark detection, both original and altered texts are converted into syntax trees, and their structures are compared for message extraction. Expanding this concept, \citet{10.1145/1178766.1178777} introduced additional syntax transformations: \textit{Activization} and \textit{Topicalization}. Moreover, research extends beyond English, with \citet{meral2009natural} analyzing 20 morphosyntactic tools in Turkish, highlighting that languages with significant suffixation and agglutination, such as Turkish, are well-suited for syntactic watermarking.

While syntactic-based watermarking effectively embeds watermarks in a concealed manner, it heavily depends on a language's grammatical rules, often requiring language-specific customization. Frequent syntactic changes in some texts may also alter their original style and fluency.

\subsection{Generation-based Watermarking}
\label{sec:generation}

The aforementioned methods have made significant strides in text watermarking. However, they often rely on specific rules that can lead to unnatural modifications, potentially degrading text quality. If these clues are detected by human attackers, they might design watermark removal attacks or attempt to forge watermarks. A groundbreaking advancement would be generating watermarked text directly from the original text and the watermark message, a technique gradually becoming feasible with the development of pretrained language models.

One approach involves designing neural networks trained to take the original text and the watermark message as inputs and output the watermarked text. \citet{abdelnabi2021adversarial} developed \texttt{AWT}, an end-to-end watermarking scheme using a transformer encoder to encode sentences and merge sentence and message embeddings. This composite is processed by a transformer decoder to generate watermarked text. For detection, the text undergoes transformer encoder layers to retrieve the secret message. Extending \texttt{AWT}, \citet{zhang2023remark} addressed disparities between dense watermarked text distributions and sparse one-hot watermark encodings with \texttt{REMARK-LLM}. This method uses a pretrained LLM for watermark insertion and introduces a reparameterization step using Gumbel-Softmax \cite{jang2016categorical} to yield sparser token distributions. A transformer-based decoder extracts messages from these embeddings. \texttt{REMARK-LLM} can embed double the signatures of \texttt{AWT} while maintaining detection efficacy, enhancing watermark payload capacity.

With LLMs' increasing capabilities in following instructions and generating high-quality text, they are becoming viable alternatives to self-designed neural networks for embedding watermarks. \citet{lau2024waterfall} introduced \texttt{WATERFALL}, which uses a watermarked LLM to paraphrase the original text, embedding a watermark while preserving semantic content. This approach combines vocabulary permutation with a novel orthogonal watermarking perturbation method to achieve high detectability and robustness. The powerful paraphrasing capabilities of LLMs enhance the naturalness of the generated text, resulting in smoother and more fluent watermarked content.

\section{Watermarking for LLMs}
\label{sec:llm}

While we've explored watermarking for existing text (\cref{sec:existing}), the rise of LLM-generated content calls for watermarking techniques during the generation process. Watermarking during text generation often yields more natural text, akin to generation-based methods (\cref{sec:generation}). This approach allows LLMs to generate watermarked text directly, which can be defined as:
\begin{equation}
\mathcal{A}(\mathbf{x}, w) = M_w(\mathbf{x}) = \mathbf{t},
\end{equation}
where $w$ is the watermark message, $x$ is the prompt, and $M_w$ is a LLM with an embedded watermark. For simplicity, we assume the watermarked text is directly generated by this LLM.





To provide a better understanding of how to add a watermark to a LLM, we first provide an overview of the process used for generating text with an LLM. Specifically, this involves three steps, LLM training, logits generation and token sampling:

\begin{itemize}[labelindent=0pt, labelsep=0.5em, leftmargin=*]
\item \textbf{Step1: LLM Training.} This step involves training a LLM, M, with a dataset D. Training objectives vary based on the application, with next token prediction being the most common \cite{radford2019language}.
\item \textbf{Step2: Logits Generation.} With the trained LLM M, given a prompt $\mathbf{x}$ and a sequence of prior tokens $\mathbf{t}^{0:(i-1)}$, the LLM predicts the next token $\mathbf{t}^{(i)}$'s probability distribution in the vocabulary $\mathcal{V}$, expressed as logits $\mathbf{l}^{(i)}$:
\begin{equation}
\mathbf{l}^{(i)} = M(\mathbf{x}, \mathbf{t}^{0:(i-1)}).
\end{equation}
\item \textbf{Step3: Token Sampling.} The next token $\mathbf{t}^{(i)}$ is selected from $\mathbf{l}^{(i)}$, using methods like nucleus sampling \cite{holtzman2019curious}, greedy decoding, or beam search. The sampling process is denoted as:
\begin{equation}
\mathbf{t}^{(i)} = S(\text{softmax}(\mathbf{l}^{(i)})).
\end{equation}
\end{itemize}
Through these steps, LLM M generates a token $\mathbf{t}^{(i)}$. For multiple tokens, logits generation and token sampling are iteratively repeated.


In aligning with the three critical phases of text generation using LLMs, watermarking techniques for LLMs are similarly categorized into three distinct types. These are: watermarking during LLM training, during logits generation, and during token sampling. Detailed discussions of these watermarking methods are presented in Section \blueref{sec:train-watermark}, \blueref{sec:logits-watermark}, and \blueref{sec:token-watermark}, respectively.

\subsection{Watermarking during Logits Generation}
\label{sec:logits-watermark}

Watermarking during logits generation refers to the insertion of a watermark message $w$ into the logits generated by LLMs. This technique, which does not require modifying the LLM parameters, is more versatile and cost-effective than training time watermarking methods.

In this context, the watermarking algorithm $\mathcal{A}$ alters the logits from the LLM to incorporate the watermark message $w$. The modified logits $ \widetilde{\mathbf{l}^{(i)}}$ can be computed as follows:
\begin{equation}
    \widetilde{\mathbf{l}^{(i)}} = \mathcal{A}(M(\mathbf{x}, \mathbf{t}^{0:(i-1)}), w) = M_w(\mathbf{x}, \mathbf{t}^{0:(i-1)}), 
\end{equation}
where $ \widetilde{\mathbf{l}^{(i)}}$ is assumed to be produced by a watermarked LLM $M_w$.

\begin{figure}[t]
\centering 
\includegraphics[width=0.98\textwidth]{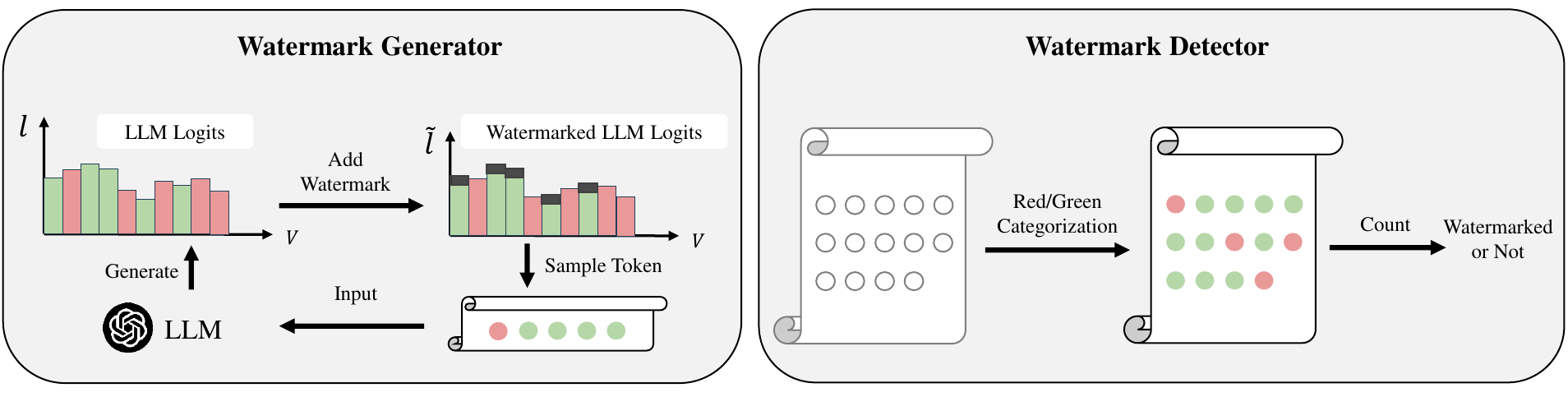} 
\caption{A more illustrative description of the KGW \cite{DBLP:conf/icml/KirchenbauerGWK23} algorithm.}
\label{fig:KGW}
\vspace{-3mm}
\end{figure}

\citet{DBLP:conf/icml/KirchenbauerGWK23} introduced the first LLM watermarking technique based on logits modification, termed KGW. This method partitions the vocabulary into a red list (R) and a green list (G) at each token position, using a hash function that depends on the preceding token. For the $i^{th}$ token generation by $M_w$, a bias $\delta$ is applied to the logits of tokens in G. The adjusted logit value, $\widetilde{\mathbf{l}_j^{(i)}}$, for a token $v_j$ at position $i$ is calculated as follows:
\begin{equation}
    \widetilde{\mathbf{l}_j^{(i)}} = M_w(\mathbf{x}, \mathbf{t}^{0:(i-1)}) = \left\{
    \begin{array}{lr}
    M(\mathbf{x}, \mathbf{t}^{0:(i-1)})[j] + \delta, &\ v_j \in G  \\
    M(\mathbf{x}, \mathbf{t}^{0:(i-1)})[j], &\ v_j \in R
    \end{array}
    \right.
    \label{KGW Soft}
\end{equation}

This algorithm biases towards green tokens, leading to a higher proportion in watermarked texts. The detector categorizes each token as red or green using the hash function and calculates the green token ratio with the z-metric, defined as:
\begin{equation}
    z = (\lvert s \rvert_{G} - \gamma T) / \sqrt{T \gamma (1 - \gamma)}
    \label{z-score}
\end{equation}
where T is the length of the text, $\gamma$ is the ratio of the green list. A text exceeding a certain green token threshold is deemed watermarked.

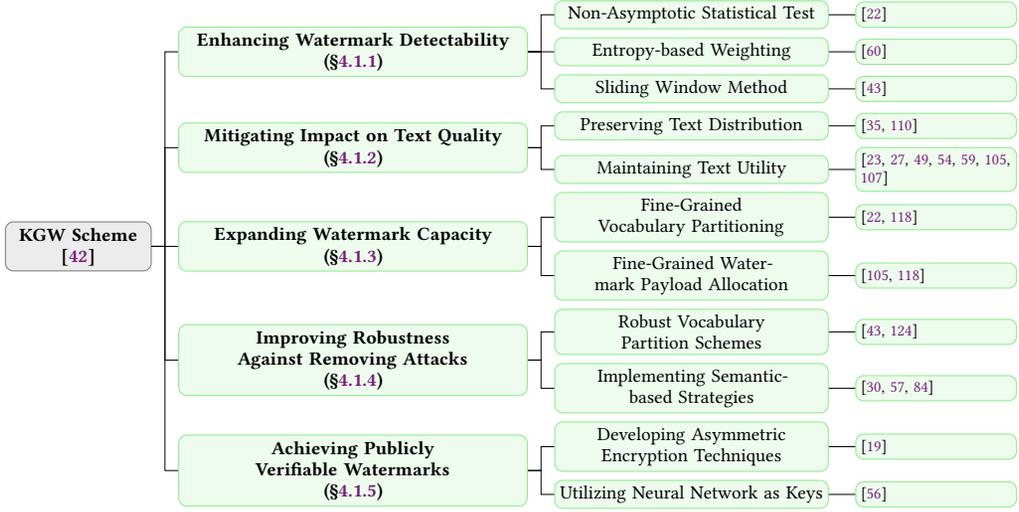
\begin{figure}[t]
\centering
\tikzset{
        my node/.style={
            draw,
            align=center,
            thin,
            text width=1.2cm, 
            rounded corners=3,
        },
        my leaf/.style={
            draw,
            align=left,
            thin,
            text width=8.5cm, 
            rounded corners=3,
        }
}
\forestset{
  every leaf node/.style={
    if n children=0{#1}{}
  },
  every tree node/.style={
    if n children=0{minimum width=1em}{#1}
  },
}
\begin{forest}
    nonleaf/.style={font=\scriptsize},
     for tree={%
        every leaf node={my leaf, font=\tiny},
        every tree node={my node, font=\tiny, l sep-=4.5pt, l-=1.pt},
        anchor=west,
        inner sep=2pt,
        l sep=10pt, 
        s sep=3pt, 
        fit=tight,
        grow'=east,
        edge={ultra thin},
        parent anchor=east,
        child anchor=west,
        if n children=0{}{nonleaf}, 
        edge path={
            \noexpand\path [draw, \forestoption{edge}] (!u.parent anchor) -- +(5pt,0) |- (.child anchor)\forestoption{edge label};
        },
        if={isodd(n_children())}{
            for children={
                if={equal(n,(n_children("!u")+1)/2)}{calign with current}{}
            }
        }{}
    }
    [\textbf{KGW Scheme \\ \cite{DBLP:conf/icml/KirchenbauerGWK23}}, draw=gray, fill=gray!15, text width=1.8cm, text=black
    [\textbf{Enhancing Watermark Detectability \\ (\cref{sec:enhance-detect})}, color=lightgreen, fill=lightgreen!15, text width=4.5cm, text=black
                [Non-Asymptotic Statistical Test, color=lightgreen, fill=lightgreen!15, text width=3.5cm, text=black            [\citep{fernandez2023three}, color=lightgreen, fill=lightgreen!15, text width=2.0cm, text=black]
                ]
                [Entropy-based Weighting,
                color=lightgreen, fill=lightgreen!15, text width=3.5cm, text=black
                [\citep{lu2024entropy}, color=lightgreen, fill=lightgreen!15, text width=2.0cm, text=black]
                ]
                [Sliding Window Method,
                color=lightgreen, fill=lightgreen!15, text width=3.5cm, text=black
                [\citep{kirchenbauer2023reliability}, color=lightgreen, fill=lightgreen!15, text width=2.0cm, text=black]
                ]
        ]
        [\textbf{Mitigating Impact on Text Quality \\ (\cref{sec:mitigate-quality})}, color=lightgreen, fill=lightgreen!15, text width=4.5cm, text=black
            [Preserving Text Distribution, color=lightgreen, fill=lightgreen!15, text width=3.5cm, text=black
                [\citep{hu2023unbiased,wu2023dipmark}, color=lightgreen, fill=lightgreen!15, text width=2.0cm, text=black]
                ]
                [Maintaining Text Utility,
                color=lightgreen, fill=lightgreen!15, text width=3.5cm, text=black
                [\citep{lee2023wrote,liu2024adaptive,wang2023towards,wouters2023optimizing,liang2024watme,guan2024codeip,fu2024watermarking},  color=lightgreen, fill=lightgreen!15, text width=2.0cm, text=black]
                ]
        ]
        [\textbf{Expanding Watermark Capacity \\ (\cref{sec:expand-capacity})}, color=lightgreen, fill=lightgreen!15, text width=4.5cm, text=black
                [Fine-Grained \\ Vocabulary Partitioning, color=lightgreen, fill=lightgreen!15, text width=3.5cm, text=black
                [\citep{fernandez2023three,yoo2023advancing}, color=lightgreen, fill=lightgreen!15, text width=2.0cm, text=black]
                ]
                [Fine-Grained Watermark Payload Allocation,
                color=lightgreen, fill=lightgreen!15, text width=3.5cm, text=black
                [\citep{wang2023towards,yoo2023advancing}, color=lightgreen, fill=lightgreen!15, text width=2.0cm, text=black]
                ]
        ]
        [\textbf{Improving Robustness Against Removing Attacks \\ (\cref{sec:remove-attack})}, color=lightgreen, fill=lightgreen!15, text width=4.5cm, text=black
                [ Robust Vocabulary Partition Schemes, color=lightgreen, fill=lightgreen!15, text width=3.5cm, text=black
                [\citep{kirchenbauer2023reliability,zhao2023provable}, color=lightgreen, fill=lightgreen!15, text width=2.0cm, text=black]
                ]
                [Implementing Semantic-based Strategies,
                color=lightgreen, fill=lightgreen!15, text width=3.5cm, text=black
                [\citep{liu2023semantic,he2024can,ren2023robust}, color=lightgreen, fill=lightgreen!15, text width=2.0cm, text=black]
                ]
        ]
        [\textbf{Achieving Publicly \\ Verifiable Watermarks \\ (\cref{sec:public-watermark})}, color=lightgreen, fill=lightgreen!15, text width=4.5cm, text=black
                [Developing Asymmetric Encryption Techniques, color=lightgreen, fill=lightgreen!15, text width=3.5cm, text=black
                [\citep{cryptoeprint:2023/1661}, color=lightgreen, fill=lightgreen!15, text width=2.0cm, text=black]
                ]
                [Utilizing Neural Network as Keys,
                color=lightgreen, fill=lightgreen!15, text width=3.5cm, text=black
                [\citep{liu2023private}, color=lightgreen, fill=lightgreen!15, text width=2.0cm, text=black]
                ]
                ]
        ],
    ]
]
\end{forest}
\caption{Taxonomy of watermarking during logits generation. The root node represents the fundamental watermarking scheme KGW \citep{DBLP:conf/icml/KirchenbauerGWK23}, followed by branches illustrating five optimizing objectives.}
\label{fig:taxonomy_of_logits}
\vspace{-3mm}
\end{figure}

KGW's detection method showed low false positive (< $3 \times 10^{-3}$\%) and false negative (< 1\%) rates in tests. Yet, real-world application challenges necessitate further optimization and design. 
The following outlines five optimization objectives, along with the improvements and explorations in watermark algorithms under each objective. The taxonomy of this section is depicted in Figure \ref{fig:taxonomy_of_logits}.

\subsubsection{Enhancing Watermark Detectability}
\label{sec:enhance-detect}
Although KGW reported high detection performance, it showed weaknesses under more rigorous detection conditions, necessitating targeted optimization of z-score computation.

One critical issue is the discrepancy between theoretical and actual false positive rates (FPRs) when the theoretical FPR is extremely low (e.g., below $10^{-6}$). This occurs because KGW assumes the z-score follows a Gaussian distribution, but this is only valid when the token length approaches infinity, and is therefore often inaccurate in practice. To resolve this, \citet{fernandez2023three} developed a non-asymptotic statistical test that adopts the accurate binomial distribution and corrects z-score calculations accordingly, aligning theoretical and actual FPRs in low-FPR scenarios.

Another challenge with KGW is its performance in low-entropy scenarios, such as code generation and machine translation. In these cases, logits vectors often display uneven distributions, reducing the impact of bias on green tokens and lowering watermark detection sensitivity. The EWD \citep{lu2024entropy} method addresses this by assigning weights to tokens based on their entropy during detection, enhancing sensitivity by emphasizing high-entropy tokens in z-score calculations.

Additionally, KGW struggles when watermarked text is mixed with extensive non-watermarked text, diluting watermark strength and affecting z-score calculations. To mitigate this, the WinMax \citep{kirchenbauer2023reliability} sliding window-based method calculates z-scores across different window sizes within the text and selects the maximum z-score as the final value, improving detection accuracy.

\subsubsection{Mitigating Impact on Text Quality}\label{sec:mitigate-quality} Watermarking can impact text quality by introducing unnatural word choices or patterns that degrade readability, coherence and text utility. An optimization perspective to mitigate this impact is to ensure that the text distribution of watermarked content remains consistent with the original output distribution of the LLM. Specifically, this means that the expected watermarked logits equal the original LLM's logits:
\begin{equation}
    \mathop{\mathbb{E}}\limits_{k\sim K}[M_w(\mathbf{x}, \mathbf{t}^{0:(i-1)})] = M(\mathbf{x}, \mathbf{t}^{0:(i-1)}), 
    \label{unbiased}
\end{equation}
where each key $k$ represents a unique red-green list split. 
This ensures that, in expectation, the watermark does not negatively affect text quality.

\citet{hu2023unbiased} noted that the KGW algorithm \cite{DBLP:conf/icml/KirchenbauerGWK23} introduces bias in its logits modification, altering the text distribution. The bias in KGW arises from applying a uniform $\delta$ to green list tokens, which disproportionately affects low-probability tokens, leading to overall bias. To counter this, \citet{hu2023unbiased} introduced two unbiased reweighting methods: $\delta$-reweight, which samples a one-hot distribution from the original logits, and $\gamma$-reweight, which halves the probability distribution range, thereby doubling the probabilities of the remaining tokens. Similarly, \citet{wu2023dipmark} proposed the $\alpha$-reweight method, which discards tokens with probabilities below $\alpha$ and adjusts the rest. These methods are theoretically unbiased, preserving the original text distribution and, consequently, maintaining text quality.

The aforementioned optimization approach is relatively theoretical. A more practical perspective focuses on mitigating the impact on text utility. One class of methods selectively applies watermarks to tokens, avoiding positions where suitable tokens are sparse. Entropy has been proposed as a criterion for watermark application, bypassing low-entropy positions \citep{lee2023wrote,wang2023towards,liu2024adaptive}. \citet{wouters2023optimizing} goes further by considering not only entropy but also the probability distribution of the red and green lists, refraining from watermarking when the probability of red tokens is too high. Another class of method is to maintain text utility by reducing the likelihood of suitable words being banned. For example, \citet{liang2024watme} proposes using mutual exclusion rules for red-green partitioning, aiming to evenly distribute semantically similar words between the red and green lists. This approach aims to prevent situations where all appropriate words are allocated to the red list. Moreover, \citet{fu2024watermarking} suggests employing semantic-aware watermarking, which involves adding words with higher relevance to the context into the green list, thus increasing the likelihood of appropriate tokens being selected. Furthermore, \citet{guan2024codeip} suggests adding an extra bias to suitable tokens, reducing the probability of their exclusion.

\subsubsection{Expanding Watermark Capacity} \label{sec:expand-capacity} The KGW watermark algorithm \cite{DBLP:conf/icml/KirchenbauerGWK23} can only verify watermark presence, classifying it as a zero-bit watermark. Yet, many applications require watermarks to convey additional information like copyright details, timestamps, or identifiers, leading to the need for multi-bit watermarks capable of extracting meaningful data. 

One possible solution is to employ fine-grained vocabulary partitioning, expanding from a binary red-green partition to a multi-color partition \citep{fernandez2023three}. To encode $b$ bits of information, the vocabulary needs to be divided into $2^b$ groups, with different watermark information reflecting preferences for tokens from different groups. Another solution is to use fine-grained watermark payload allocation \citep{wang2023towards}, dividing the text into multiple chunks, with each chunk encoding a portion of the bit information. However, if either of these approaches is too fine-grained in its partitioning, the encoding strength may be insufficient, leading to reduced watermark extraction accuracy. To mitigate this issue, \citet{yoo2023advancing} attempts to combine these two methods, ensuring an adequate watermark capacity while avoiding overly fine-grained partitioning of any kind, and aims to find an optimal balance of granularity through experimentation.

\subsubsection{Improving Robustness Against Removing Attacks} \label{sec:remove-attack}
As discussed in Section \ref{sec:pre}, an effective watermarking algorithm must be robust against removal attacks, ensuring the watermark remains detectable. These attacks usually modify the text without altering its semantic content. While the KGW algorithm \cite{DBLP:conf/icml/KirchenbauerGWK23} exhibited some robustness in their experiments, there is still room for improvement. 

One optimization approach is to adopt more robust vocabulary partition schemes. The original KGW \citep{DBLP:conf/icml/KirchenbauerGWK23} algorithm utilizes all token information within the preceding context window to map to a hash value for red-green partitioning. \citet{kirchenbauer2023reliability} further elaborated on more robust partition strategies, such as using only the samllest token id in the preceding context window for hashing, which is more resilient to text editing. Additionally, \citet{zhao2023provable} proved that a fixed global split of red and green lists offers greater resistance to removal attacks.

As watermark removal attacks usually preserve the semantic content of the text, several studies have developed methods to integrate semantic information into the design of watermarking algorithms. For example, \citet{liu2023semantic} trained a watermarked LLM that directly converts text semantics into red-green partitions, ensuring that similar text semantics result in similar partition outcomes, thereby achieving robustness. \citet{he2024can} improves the watermark LLM by adding constraints that ensure semantically similar tokens fall into the same color list, thereby further enhancing robustness. \citet{ren2023robust} converted semantic embeddings into semantic values through weighted embedding pooling followed by discretizing using NE-Ring, and then divided the vocabulary into red-list and green-list based on these semantic values.

\subsubsection{Achieving Publicly Verifiable Watermarks} \label{sec:public-watermark} Achieving publicly verifiable watermarks is significant as it allows anyone to authenticate the origin and integrity of the content without requiring access to a secret key. This reduces service overhead in private detection scenarios (where the detector is placed behind an API) and its transparency also enhances trust and accountability. Most previous watermarking algorithms cannot achieve public verifiability because their watermark generation details are involved in detection (e.g., the hash key in KGW). As a result, exposing the detector also means exposing the generator, making the watermark vulnerable to targeted removal or forgery.

To achieve publicly verifiable watermarks, \citet{cryptoeprint:2023/1661} have utilized digital signature technology from the field of cryptography. This approach involves generating watermarks using a private key and verifying them with a public key. However, verification via a public key relies on features extracted from the text, which can still be exploited to some extent to forge watermarks. Further advancing this field, \citet{liu2023private}  proposed the use of different neural networks for watermark generation and detection. Due to the black-box nature of neural networks, the details of watermark generation are not exposed, which could defend against watermark forgeries in public detection scenarios.

\subsection{Watermarking during Token Sampling}
\label{sec:token-watermark}

\begin{figure}[t]
\centering
\tikzset{
        my node/.style={
            draw,
            align=center,
            thin,
            text width=1.2cm, 
            rounded corners=3,
        },
        my leaf/.style={
            draw,
            align=left,
            thin,
            text width=8.5cm, 
            rounded corners=3,
        }
}
\forestset{
  every leaf node/.style={
    if n children=0{#1}{}
  },
  every tree node/.style={
    if n children=0{minimum width=1em}{#1}
  },
}
\begin{forest}
    nonleaf/.style={font=\scriptsize},
     for tree={%
        every leaf node={my leaf, font=\tiny},
        every tree node={my node, font=\tiny, l sep-=4.5pt, l-=1.pt},
        anchor=west,
        inner sep=2pt,
        l sep=10pt, 
        s sep=3pt, 
        fit=tight,
        grow'=east,
        edge={ultra thin},
        parent anchor=east,
        child anchor=west,
        if n children=0{}{nonleaf}, 
        edge path={
            \noexpand\path [draw, \forestoption{edge}] (!u.parent anchor) -- +(5pt,0) |- (.child anchor)\forestoption{edge label};
        },
        if={isodd(n_children())}{
            for children={
                if={equal(n,(n_children("!u")+1)/2)}{calign with current}{}
            }
        }{}
    }
    [\textbf{Watermarking During \\ Token Sampling}, draw=harvestgold, fill=harvestgold!15, text width=4cm, text=black
        [\textbf{Token-level Sampling Watermarking (\cref{sec:token-level sampling})}, color=harvestgold, fill=harvestgold!15, text width=5.5cm, text=black
                [\citep{christ2024undetectable,aronsonpowerpoint,kuditipudi2023robust}, color=harvestgold, fill=harvestgold!15, text width=0.9cm, text=black
        ]
        ]
        [\textbf{Sentence-level Sampling Watermarking (\cref{sec:sentence-level sampling})}, color=harvestgold, fill=harvestgold!15, text width=5.5cm, text=black
            [\citep{hou2023semstamp,hou2024k}, color=harvestgold, fill=harvestgold!15, text width=0.9cm, text=black
        ]
    ]
]
]
\end{forest}
\caption{Taxonomy of watermarking during token sampling.}
\label{fig:taxonomy_of_sampling}
\vspace{-3mm}
\end{figure}
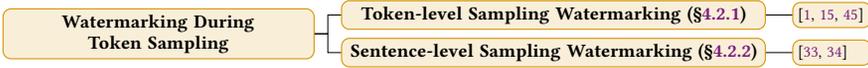

The previous section primarily focused on incorporating watermarks during the logits generation phase for LLMs. 
In this section, we will introduce a technique of watermarking during token sampling, which does not alter the logits but utilize watermark message to guide the sampling process. Based on the granularity of guiding, this technique can be divided into two main approaches (as depicted in Figure \ref{fig:taxonomy_of_sampling}): token-level sampling watermarking (\cref{sec:token-level sampling}), which embeds watermarks during each token's sampling, and sentence-level sampling watermarking (\cref{sec:sentence-level sampling}), which uses watermark message to guide the sampling of entire sentences.

\subsubsection{Token-level Sampling Watermarking}
\label{sec:token-level sampling}
The principle of incorporating watermarks during the token sampling phase is derived from the randomness inherent in token sampling. In this scenario, watermarks can be introduced using a fixed random seed, where a pseudo-random number generator produces a sequence of pseudo-random numbers to guide the sampling of each token. For watermark detection, it is only necessary to assess the alignment between the text tokens and the pseudo-random numbers, specifically evaluating whether the choice of each token in the text matches with the corresponding value in the random number sequence. 

For instance, \citet{christ2024undetectable} proposed a watermarking algorithm designed for a toy LLM with a vocabulary consisting of only the digits 0 and 1, with the pseudo-random numbers represented as a series of values $u \in [0,1]$. If the predicted probability for a certain position exceeds the corresponding pseudo-random number, then 1 is sampled at that position, otherwise 0. In the detection of watermarks, it can be determined whether the values of the pseudo-random numbers corresponding to the positions with 1 in the binary tokens are significantly higher than those with 0. Around the same time, \citet{aronsonpowerpoint} proposed a watermarking algorithm based on a similar idea, but suitable for real LLMs. In this case, the LLM (with a vocabulary size of \(|V|\)) outputs a probability vector \( p_i = (p_{i1}, ..., p_{i|V|}) \) at position \( i \), and the pseudo-random sequence at position \( i \) is also transformed from a single number (0 or 1) into a pseudo-random vector \( r_i = (r_{i1}, ..., r_{i|V|}) \). During sampling, exp-minimum sampling is applied to choose the token $j$ that maximizes \( r_{ij}^{1/p_{ij}}\). To check a watermark, the alignment between the text and the pseudo-random vector sequence is assessed to see if it exceeds a threshold. 

However, these methods still faces two challenges: 1) the detection algorithm is not robust enough against watermark removal attacks, which involves certain text modifications, and 2) due to the fixed nature of pseudo-random numbers, the LLM with watermark will generate the same text for the same prompt each time, thereby losing the inherent randomness in text generation by LLM. To address these issues, \citet{kuditipudi2023robust} proposed the use of a pseudo-random number sequence significantly longer than the text, randomly selecting a starting position from the sequence for each watermark insertion to introduce randomness. Additionally, during watermark detection, they incorporate a soft notion of edit distance (i.e., Levenshtein distance) into the computation of the alignment between text and the pseudo-random number sequence. This approach significantly enhances the robustness of the watermarking algorithm against watermark removal attacks. 

\subsubsection{Sentence-level Sampling Watermarking}
\label{sec:sentence-level sampling}
Token-level sampling watermarking algorithms may not be robust to token-level text edits. However, since such edits often don't significantly change sentence semantics, some methods leverage sentence-level sampling watermarking to achieve better robustness.

Building on this idea, SemStamp \citep{hou2023semstamp} partitions the semantic embedding space into a watermarked region and a non-watermarked region. The algorithm performs sentence-level rejection sampling until the sampled sentence falls within the watermarked region. SemStamp employs Locality-Sensitive Hashing to randomly partition these regions, which can result in semantically similar sentences being placed into different regions, thereby diminishing the robustness of the watermarking algorithm. To address this issue, $k$-SemStamp \citep{hou2024k} utilizes $k$-means clustering to divide the semantic space into $k$ regions, ensuring that semantically similar sentences are grouped into the same region. Each region is then designated as either a watermarked region or a non-watermarked region. This approach ensures that a sentence remains within the same region even after semantic-preserving modifications, thereby enhancing robustness against semantic-invariant text editing attacks.

Current research in sampling-based watermarking is limited, indicating room for advancement. The effectiveness and robustness of these methods warrant further exploration through experiments and real-world applications.

\subsection{Watermarking during LLM Training}  
\label{sec:train-watermark}

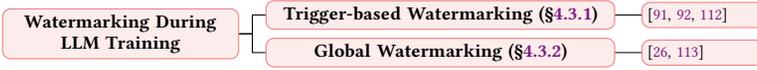
\begin{figure}[t]
\centering
\tikzset{
        my node/.style={
            draw,
            align=center,
            thin,
            text width=1.2cm, 
            rounded corners=3,
        },
        my leaf/.style={
            draw,
            align=left,
            thin,
            text width=8.5cm, 
            rounded corners=3,
        }
}
\forestset{
  every leaf node/.style={
    if n children=0{#1}{}
  },
  every tree node/.style={
    if n children=0{minimum width=1em}{#1}
  },
}
\begin{forest}
    nonleaf/.style={font=\scriptsize},
     for tree={%
        every leaf node={my leaf, font=\tiny},
        every tree node={my node, font=\tiny, l sep-=4.5pt, l-=1.pt},
        anchor=west,
        inner sep=2pt,
        l sep=10pt, 
        s sep=3pt, 
        fit=tight,
        grow'=east,
        edge={ultra thin},
        parent anchor=east,
        child anchor=west,
        if n children=0{}{nonleaf}, 
        edge path={
            \noexpand\path [draw, \forestoption{edge}] (!u.parent anchor) -- +(5pt,0) |- (.child anchor)\forestoption{edge label};
        },
        if={isodd(n_children())}{
            for children={
                if={equal(n,(n_children("!u")+1)/2)}{calign with current}{}
            }
        }{}
    }
    [\textbf{Watermarking During \\ LLM Training} , draw=lightcoral, fill=lightcoral!15, text width=3cm, text=black
        [\textbf{Trigger-based Watermarking (\cref{sec:trigger-based})}, color=lightcoral, fill=lightcoral!15, text width=4.5cm, text=black
                [\citep{sun2022coprotector, sun2023codemark,xu2024hufu} , color=lightcoral, fill=lightcoral!15, text width=1.5cm, text=black
        ]
        ]
        [\textbf{Global Watermarking (\cref{sec:global})}, color=lightcoral, fill=lightcoral!15, text width=4.5cm, text=black
            [\citep{gu2024on,xu2024learning}, color=lightcoral, fill=lightcoral!15, text width=1.5cm, text=black
        ]
    ]
]
]
\end{forest}
\caption{Taxonomy of watermarking during LLM training.}
\label{fig:taxonomy_of_training}
\vspace{-5mm}
\end{figure}

Although adding watermarks during the logits generation (\cref{sec:logits-watermark}) and token sampling (\cref{sec:token-watermark}) stages can be effective during inference, they are not suitable for open-source LLMs. This is because watermarking code added after the logits output can be easily removed. Therefore, for open-source LLMs, watermarks must be embedded into the LLM's parameters during training. As shown in Figure \ref{fig:taxonomy_of_training}, training-time watermarking can be categorized into trigger-based watermarks, which are effective only for specific inputs, and global watermarks, which are intended to work for all inputs.

\subsubsection{Trigger-based Watermarking} \label{sec:trigger-based} Trigger-based Watermarking is a type of backdoor watermarking that introduces specific triggers into an LLM. When these triggers appear in the input, the LLM exhibits specific behaviors (specific formats or outputs). Such watermarks can be added by dataset providers to protect dataset copyright, or by LLM providers to protect LLM copyright.

\citet{sun2022coprotector} proposed CoProtector for the code generation task, using word-level or sentence-level modifications in the code as triggers to generate corrupting code, which typically has incorrect functionality. Further, \citet{sun2023codemark} proposed CodeMark, which uses semantically invariant code transformations as triggers, ensuring the correct functionality of the code while embedding a trigger-based watermark with minimal impact on LLM performance.

In the context of protecting LLM copyright, \citet{xu2024hufu} proposed the Hufu watermark. This watermark does not rely on specific input triggers but uses a particular input format as a trigger. It leverages the permutation equivariance property of transformers, training the LLM to recognize a specific permutation as a watermark.


\subsubsection{Global Watermarking} \label{sec:global} Although trigger-based text watermarking is effective in many cases, it only works when specific triggers are present and cannot work for all inputs. 

Global watermarks can add detectable markers to all content generated by LLMs, enabling content tracking. \citet{gu2024on} explored the learnability of watermarks, investigating whether LLMs can directly learn to generate watermarked text. They proposed two learning methods: sampling-based watermark distillation and logit-based watermark distillation. These methods offer the possibility of transforming inference-time watermarking into inherent LLM parameters. \citet{xu2024learning} further proposed using reinforcement learning to optimize LLM watermarks. They used reinforcement learning techniques to optimize LLMs based on feedback from watermark detectors, embedding watermarks into the LLM. Experimental results show that this method achieves near-perfect watermark detection and strong resistance to interference, significantly improving watermark effectiveness and robustness. However, due to the black-box nature of LLMs, this watermark training method may be less stable on out-of-distribution data compared to inference-time watermarks.

\section{Evaluation Metrics for Text Watermarking}
\label{sec:evaluation}

\begin{figure}[t]
\centering
\tikzset{
        my node/.style={
            draw,
            align=center,
            thin,
            text width=1.2cm, 
            rounded corners=3,
        },
        my leaf/.style={
            draw,
            align=left,
            thin,
            text width=8.5cm, 
            rounded corners=3,
        }
}
\forestset{
  every leaf node/.style={
    if n children=0{#1}{}
  },
  every tree node/.style={
    if n children=0{minimum width=1em}{#1}
  },
}
\begin{forest}
    nonleaf/.style={font=\scriptsize},
     for tree={%
        every leaf node={my leaf, font=\scriptsize},
        every tree node={my node, font=\scriptsize, l sep-=4.5pt, l-=1.pt},
        anchor=west,
        inner sep=2pt,
        l sep=10pt, 
        s sep=3pt, 
        fit=tight,
        grow'=east,
        edge={ultra thin},
        parent anchor=east,
        child anchor=west,
        if n children=0{}{nonleaf}, 
        edge path={
            \noexpand\path [draw, \forestoption{edge}] (!u.parent anchor) -- +(5pt,0) |- (.child anchor)\forestoption{edge label};
        },
        if={isodd(n_children())}{
            for children={
                if={equal(n,(n_children("!u")+1)/2)}{calign with current}{}
            }
        }{}
    }
    [\textbf{Evaluation Metrics}, draw=gray, fill=gray!15, text width=1.5cm, text=black
    [\textbf{Detectability} \\ ({\cref{sec:detectability}}), color=brightlavender, fill=brightlavender!15, text width=2cm, text=black
            [Zero-bit Watermark (\cref{sec:zero-bit}), color=brightlavender, fill=brightlavender!15, text width=3.0cm, text=black
                [{z-score, p-value, binary classification metrics}, color=brightlavender, fill=brightlavender!15, text width=5.8cm, text=black]
            ]
            [Multi-bit Watermark (\cref{sec:multi-bit}), color=brightlavender, fill=brightlavender!15, text width=3.0cm, text=black
                [ {BER, bit accuracy, payload}, color=brightlavender, fill=brightlavender!15, text width=5.8cm, text=black]
            ],
            [Watermark Size (\cref{sec:eval-size}), color=brightlavender, fill=brightlavender!15, text width=3.0cm, text=black
            ],
        ]
        [\textbf{Quality Impact}, color=lightgreen, fill=lightgreen!15, text width=2cm, text=black
            [Quality Impact of \\ Watermarked Text (\cref{sec:quality-text}), color=lightgreen, fill=lightgreen!15, text width=3.0cm, text=black
                [Comparative Metrics (\cref{sec:eval-text-comp}), color=lightgreen, fill=lightgreen!15, text width=2.8cm, text=black
                    [{Meteor, BLEU, semantic score, entailment score}, color=lightgreen, fill=lightgreen!15, text width=2.5cm, text=black]
                ],
                [Single Metrics (\cref{sec:eval-text-single}), color=lightgreen, fill=lightgreen!15, text width=2.8cm, text=black
                [{PPL, human evaluation}, color=lightgreen, fill=lightgreen!15, text width=2.5cm, text=black]
                ]
            ],
            [Output Performance of \\Watermarked LLM (\cref{sec:quality-llm}), color=lightgreen, fill=lightgreen!15, text width=3.0cm, text=black
                [{text completion, code generation, machine translation, text summarization, question answering, math reasoning, knowledge probing, instruction following}, color=lightgreen, fill=lightgreen!15, text width=5.8cm, text=black]
            ],
            [Output Diversity of \\Watermarked LLM (\cref{sec:quality-diverse}), color=lightgreen, fill=lightgreen!15, text width=3.0cm, text=black
                [{Seq-Rep-N, log diversity, Ent-3, Sem-Ent}, color=lightgreen, fill=lightgreen!15, text width=5.8cm, text=black]
            ],
        ],
        [\textbf{Robustness \\ under \\ Watermark Attack}, color=harvestgold, fill=harvestgold!15, text width=2cm, text=black
            [Untargeted Attack (\cref{sec:untargeted-attack}), color=harvestgold, fill=harvestgold!15, text width=3.0cm, text=black
                [Character-level (\cref{sec:eval-unt-char}), color=harvestgold, fill=harvestgold!15, text width=2.8cm, text=black
                    [{homoglyph attack}, color=harvestgold, fill=harvestgold!15, text width=2.5cm, text=black]
                ],
                [{Word-level (\cref{sec:eval-unt-word-exist},\cref{sec:eval-unt-word-gen})}, color=harvestgold, fill=harvestgold!15, text width=2.8cm, text=black
                [{word insertion, word deletion, word replacement, emoji attack}, color=harvestgold, fill=harvestgold!15, text width=2.5cm, text=black]
                ],
                [{Doc-level (\cref{sec:eval-unt-para},\cref{sec:eval-unt-cp},\cref{sec:other-doc})}, color=harvestgold, fill=harvestgold!15, text width=2.8cm, text=black
                [{back-translation, paraphrasing, copy-paste attack, syntax transformation, CWRA, watermark collision}, color=harvestgold, fill=harvestgold!15, text width=2.5cm, text=black]
                ]
                [{Fine-tuning Attack (\cref{sec:fine-tuning-attack})}, color=harvestgold, fill=harvestgold!15, text width=2.8cm, text=black]
            ],
            [Targeted Attack (\cref{sec:targeted-attack}), color=harvestgold, fill=harvestgold!15, text width=3.0cm, text=black
                [Targeted Attack Against KGW (\cref{sec:eval-t-kgw}), color=harvestgold, fill=harvestgold!15, text width=2.8cm, text=black
                    [{Spoofing attack, SCTS, MIP, WS}, color=harvestgold, fill=harvestgold!15, text width=2.5cm, text=black]
                ],
                [{Watermark Distillation (\cref{sec:eval-t-distill})}, color=harvestgold, fill=harvestgold!15, text width=3.2cm, text=black]
            ],
        ]
    ]
    \end{forest}
\caption{Taxonomy of Evaluation Metrics for Text Watermarking.}
\label{fig_taxonomy_of_metrics}
\vspace{-5mm}
\end{figure}
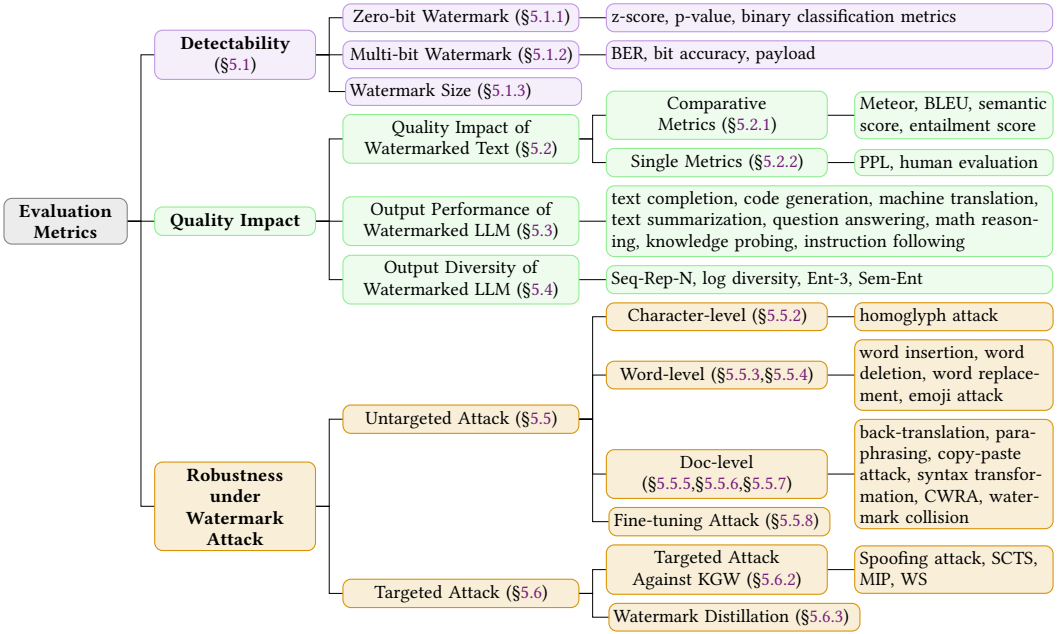

In Sections \blueref{sec:existing} and \blueref{sec:llm}, we provided a comprehensive overview of existing text watermarking techniques. A thorough evaluation of text watermarking algorithms is crucial. As illustrated in Figure \ref{fig_taxonomy_of_metrics}, this section details the evaluation metrics from multiple perspectives: (1) the detectability of watermarking algorithms (\cref{sec:detectability}), (2) the impact of watermarking on the quality of targeted texts (\cref{sec:quality-text}) and LLMs (\cref{sec:quality-llm} and \cref{sec:quality-diverse}), and (3) the robustness of watermarking algorithms against untargeted (\cref{sec:untargeted-attack}) and targeted watermark attacks (\cref{sec:targeted-attack}). In addition, we enumerate representative evaluation benchmarks and tools (\cref{sec:benchmark}). 


\subsection{Detectability}

\label{sec:detectability}

For text watermarking algorithms, the basic requirement is that the watermarked text can be detected. In this section, we will summarize how watermarking algorithms measure their detectability. We will introduce the detection metrics for zero-shot and multi-bit watermarking algorithms, as well as the watermark size, which indicates how long the text needs to be for detection.



\subsubsection{Zero-bit Watermark}  
\label{sec:zero-bit}
In zero-bit watermarking, the goal is to detect the presence of a watermark. Current watermark algorithms typically provide a detector that uses hypothesis testing to generate a z-score or p-value \cite{DBLP:conf/icml/KirchenbauerGWK23, kuditipudi2023robust, liu2023private}, along with a threshold to distinguish whether a text contains a watermark. For testing, a dataset with an equal number of watermarked and human texts is usually constructed. The detector is then used to evaluate this dataset, calculating the F1 score and corresponding false positive and false negative rates. The false positive rate, which indicates the probability of misclassifying human text as watermarked, is particularly important as it can have more severe consequences than false negatives.

The challenge with this detection method lies in selecting an appropriate threshold, as different methods may have different threshold selection approaches. For example, some studies \cite{DBLP:conf/icml/KirchenbauerGWK23, zhao2023provable, liu2023private, liu2023semantic} report F1 scores at fixed false positive rates of 1\% and 10\%, while others \cite{liu2023semantic} show the best F1 scores across all thresholds to facilitate a fairer comparison of algorithm performance.

\subsubsection{Multi-bit Watermark}  \label{sec:multi-bit} 
In multi-bit watermarking methods \cite{wang2023towards, rizzo2016content, abdelnabi2021adversarial, yang2022tracing, yoo2023robust, yoo2023advancing}, the watermark detection algorithm must not only detect the presence of a watermark but also extract specific information. For example, a watermarked text might encode specific data like \textit{"This text is generated by GPT-4 on June 6 by the Administrator"} \cite{wang2023towards}. Common detection metrics include bit error rate (BER) \cite{yoo2023robust} and bit accuracy \cite{yoo2023advancing, abdelnabi2021adversarial}. For a watermark message $w$ encoded as $n$ bits, represented as $w = b_1b_2…b_{n}$, where each $b_i$ is binary, BER refers to the probability of incorrectly predicted bits, while bit accuracy refers to the proportion of correctly predicted bits.

Additionally, the bit capacity or payload of a watermark algorithm is a key evaluation metric, typically referred to as Bits Per Watermark \cite{yang2022tracing, yoo2023robust} or code rate \cite{wang2023towards, rizzo2016content}. Payload is calculated by dividing the total number of bits of the watermark information by the number of tokens. 



\subsubsection{Watermark Size} 
\label{sec:eval-size} 
Generally, for a text watermarking algorithm, the longer the text, the easier it is for the watermark to be detected because longer texts provide more modification space. Thus, determining how long a text needs to be for the watermark to be reliably detected becomes a crucial metric, known as the watermark size.
\citet{piet2023mark} studied the watermark size of current mainstream watermarking algorithms, exploring the minimum length required to achieve a 2\% false positive rate. The study found that among the KGW \cite{DBLP:conf/icml/KirchenbauerGWK23}, Aar \cite{aronsonpowerpoint}, and KTH \cite{kuditipudi2023robust} algorithms, KGW requires the shortest detection length, indicating that KGW has the best watermark size. Currently, there is still limited research on watermark size, and future work is recommended to include watermark size as an evaluation metric.

Typically, for a text watermarking algorithm, high detectability is a relatively low requirement.  More importantly, these algorithms should have minimal impact on text quality and high robustness against various attacks. In the following two sections, we will separately introduce how to evaluate the quality of watermarked texts (\cref{sec:quality-text}) and the quality assessment of watermarked LLMs (\cref{sec:quality-llm}).


\subsection{Quality Impact of Watermarked Text}



\begin{table}[t]
\centering
\caption{The quality evaluation metrics for different text watermarking algorithms regarding their impact on text or LLM quality. \textit{w. Ext. LLM} indicates whether external LLMs are used, \textit{Diverse Eval} indicates whether text diversity is evaluated, \textit{Time Cmplx.} indicates the time complexity of the evaluation, \textit{Eval Type} indicates whether the evaluation scores single texts (\textit{Single}) or compares pairs of texts (\textit{Comp.}), \textit{Tested Algorithms} lists the algorithms tested on each metric, and \textit{Quality Preserve} refers to algorithms that demonstrate the preservation of text quality under the corresponding metric, based on the analysis provided in the respective studies conducting the tests.}
\resizebox{\textwidth}{!}{%
\begin{tabular}{>{\raggedright\arraybackslash}m{3.5cm}@{} 
>{\centering\arraybackslash}m{3cm}
>{\centering\arraybackslash}m{1.5cm}@{} 
>{\centering\arraybackslash}m{1.5cm}@{} 
>{\centering\arraybackslash}m{1.2cm}@{} 
>{\centering\arraybackslash}m{1.2cm} 
>{\centering\arraybackslash}m{4cm}
>{\centering\arraybackslash}m{3cm}
}
\toprule
\multicolumn{1}{l}{\textbf{Evaluation Task}} & \makecell{\textbf{Metric}} & \makecell{\textbf{w. Ext.} \\ \textbf{LLM?}} & \makecell{\textbf{Diversity} \\ \textbf{Eval?}} & \makecell{\textbf{Time} \\ \textbf{Cmplx.}} & \makecell{\textbf{Eval} \\ \textbf{Type}} & \makecell{\textbf{Tested} \\ \textbf{Algorithms}} & \makecell{\textbf{Quality} \\ \textbf{Preserve}} \\
\midrule
\rowcolor{green!10} \multicolumn{8}{c}{\textbf{Quality Evaluation for Watermarked Text}} 
\\ \midrule
\diagbox[width=9em,height=1em]{}{}& Semantic Score & \ding{51} & \ding{55} & M. & 	Comp. & \cite{munyer2023deeptextmark, yoo2023robust, abdelnabi2021adversarial, yang2023watermarking, zhang2023remark, yang2022tracing} & \cite{abdelnabi2021adversarial,munyer2023deeptextmark,yang2023watermarking,yang2022tracing,yoo2023robust,zhang2023remark}\\ 
\diagbox[width=9em,height=1em]{}{}  & Meteor Score & \ding{55} & \ding{55} & M. & 	Comp. & \cite{yang2023watermarking, abdelnabi2021adversarial} & \cite{abdelnabi2021adversarial}\\ 
\diagbox[width=9em,height=1em]{}{}  & Entailment Score & \ding{51} & \ding{55} & M. & 	Comp. & \cite{yoo2023robust, yang2022tracing} & \cite{yang2022tracing,yoo2023robust}\\
\diagbox[width=9em,height=1em]{}{}  & BLEU Score & \ding{55} & \ding{55} & L. & 	Comp. & \cite{zhang2023remark, 10.1145/1178766.1178777,sato2023embarrassingly} & \cite{zhang2023remark,10.1145/1178766.1178777,sato2023embarrassingly}\\
\diagbox[width=9em,height=1em]{}{}  & Perplexity & \ding{51} & \ding{55} & M. & 	Single & \cite{sato2023embarrassingly} & \cite{sato2023embarrassingly}\\
\diagbox[width=9em,height=1em]{}{}  & Human Evaluation & \ding{55} & \ding{55} & H. & 	Single & \cite{yoo2023robust, abdelnabi2021adversarial} & \cite{abdelnabi2021adversarial,yoo2023robust}\\ 
  \midrule
  \rowcolor{blue!10} \multicolumn{8}{c}{\textbf{Quality Evaluation for Watermarked LLM}} 
\\ \midrule
\multirow{7}{*}{\makecell[{{p{3.5cm}}}]{\textbf{Text Completion}}} 
&   PPL &   \ding{51} &  \ding{55} &   M. &   Single &   \cite{DBLP:conf/icml/KirchenbauerGWK23, liu2023private, liu2023semantic, zhao2023provable, liang2024watme, liu2024adaptive, lu2024entropy, ren2023robust, yoo2023advancing, wang2023towards, wouters2023optimizing, hou2023semstamp, hou2024k, gu2024on} & \cite{hou2023semstamp,hou2024k,DBLP:conf/icml/KirchenbauerGWK23,liang2024watme,liu2023private,liu2023semantic,lai2024adaptive,lu2024entropy,ren2023robust,wang2023towards,wouters2023optimizing,yoo2023advancing,zhao2023provable} \\
  &   P-SP &  \ding{51} &   \ding{55} &  M. &  	Comp. &  \cite{yoo2023advancing} & \cite{yoo2023advancing}\\
  &   GPT4-Score &  \ding{51} &   \ding{55} &  M. &  	Single &  \cite{cryptoeprint:2023/1661} & \cite{cryptoeprint:2023/1661}\\
  & Seq-Rep-N & \ding{55} & \ding{51} & L. &  Single & \cite{gu2024on} & \cite{gu2024on}\\
  &  Log Div. &  \ding{55} &  \ding{51} &  L. &  	Single &  \cite{kirchenbauer2023reliability} & \cite{kirchenbauer2023reliability}\\
    &  Ent-3 &  \ding{55} &  \ding{51} &  L. &  	Single &  \cite{hou2023semstamp, hou2024k} & \cite{hou2023semstamp,hou2024k}\\
      &  Sem-Ent &  \ding{51} &  \ding{51} &  M. &  	Single &  \cite{hou2023semstamp,hou2024k} & \cite{hou2023semstamp,hou2024k}\\
 \midrule
\multirow{3}{*}{\makecell[{{p{3.5cm}}}]{\textbf{Code Generation}}} 
&\centering  Pass@k & \ding{55} &  \ding{55}& L. & Single & \cite{fernandez2023three, lee2023wrote, lu2024entropy} & \cite{fernandez2023three,lu2024entropy,lee2023wrote} \\
&\centering  CodeBlue & \ding{55} &  \ding{55} & L. & Comp. & \cite{guan2024codeip} & \cite{guan2024codeip}\\
&\centering  Edit Sim & \ding{55} &  \ding{55} & L. & Comp. & \cite{DBLP:conf/icml/KirchenbauerGWK23,zhao2023provable,kirchenbauer2023reliability} & -\\
 \midrule
\multirow{3}{*}{\makecell[{{p{3.5cm}}}]{\textbf{Machine  \newline Translation}}} 
&\centering  BLEU & \ding{55} &  \ding{55} & L. & Comp. & \cite{hu2023unbiased, wu2023dipmark, liu2023private} & \cite{hu2023unbiased,liu2023private,wu2023dipmark} \\
&\centering  BERTScore & \ding{51} &  \ding{55} & M. & Comp. & \cite{hu2023unbiased, wu2023dipmark} & \cite{hu2023unbiased,wu2023dipmark}\\
&\centering  PPL & \ding{51} &  \ding{55} & M. & Single & \cite{hu2023unbiased, wu2023dipmark} & \cite{hu2023unbiased,wu2023dipmark}\\
 \midrule
 \multirow{3}{*}{\makecell[{{p{3.5cm}}}]{\textbf{Text  \newline Summarization}}} 
&\centering  BLEU & \ding{55} &  \ding{55} & L. & Comp. & \cite{hu2023unbiased, wu2023dipmark, DBLP:conf/icml/KirchenbauerGWK23,zhao2023provable,kirchenbauer2023reliability, he2024can} & \cite{hu2023unbiased,wu2023dipmark}\\
&\centering  BERTScore & \ding{51} &  \ding{55} & M. & Comp. & \cite{hu2023unbiased, wu2023dipmark} & \cite{hu2023unbiased,wu2023dipmark} \\
&\centering  PPL & \ding{51} & \ding{55} & M. & Single & \cite{hu2023unbiased, wu2023dipmark} & \cite{hu2023unbiased,wu2023dipmark}\\
 \midrule
  \multirow{4}{*}{\makecell[{{p{3.5cm}}}]{\textbf{Question  \newline Answering}}} 
&\centering  ROUGE & \ding{55} &  \ding{55} & L. & Comp. & \cite{DBLP:conf/icml/KirchenbauerGWK23,zhao2023provable,kirchenbauer2023reliability} & - \\
&\centering  Exact Match & \ding{55} &  \ding{55} & L. & Comp. & \cite{fernandez2023three} & \cite{fernandez2023three}\\
&\centering  GPT-Truth & \ding{51} & \ding{55} & M. & Single & \cite{liang2024watme} & \cite{liang2024watme} \\
&\centering  GPT-Info & \ding{51} &  \ding{55}& M. & Single & \cite{liang2024watme} & \cite{liang2024watme}\\
 \midrule
\textbf{Math Reasoning} & Accuracy & \ding{55} &  \ding{55} & L. & Comp. & \cite{fernandez2023three, liang2024watme} & \cite{liang2024watme,fernandez2023three}\\  \midrule
\textbf{Knowledge Probing} & F1 Score & \ding{55} &  \ding{55} & L. & Comp. & \cite{DBLP:conf/icml/KirchenbauerGWK23,zhao2023provable,kirchenbauer2023reliability} & - \\  \midrule
\textbf{ Instruction Following} & GPT4-Judge & \ding{51} &  \ding{55} & M. & Comp. & \cite{DBLP:conf/icml/KirchenbauerGWK23,zhao2023provable,kirchenbauer2023reliability} & - \\
\bottomrule
\end{tabular}
}

\label{tab:quality_metrics}
\vspace{-3mm}
\end{table}


\label{sec:quality-text} The quality evaluation of watermarked text primarily targets the series of algorithms for watermarking existing text (\cref{sec:existing}). In these algorithms, the input is a non-watermarked text, and the output is the modified watermarked text. Therefore, the key to evaluating the quality of watermarked text is comparing the quality differences between the watermarked text and the original text.
There are two evaluation methods: one uses comparative metrics such as semantic score and BLEU score \cite{munyer2023deeptextmark, yoo2023robust, abdelnabi2021adversarial, yang2023watermarking, zhang2023remark, yang2022tracing}; the other involves scoring the original text and the watermarked text separately and then comparing the scores \cite{sato2023embarrassingly}.

\subsubsection{Comparative evaluation metrics}
\label{sec:eval-text-comp}
For comparative evaluation metrics, the main purpose is to assess the similarity between watermarked text and the original text. Based on the method of evaluating similarity, these metrics can be divided into two categories: surface feature-based metrics, such as Meteor Score \cite{banerjee2005meteor} and BLEU Score\cite{papineni2002bleu}, and semantic feature-based metrics, such as Semantic Score and Entailment Score.

Meteor score and BLEU score are important evaluation metrics in the field of machine translation, and when applied to text watermarking, the original text can be used as the reference text. BLEU focuses on the n-gram overlap between the target (watermarked) text and the reference text, providing a composite score by calculating precision and length penalty. However, BLEU's limitation is its overemphasis on exact matches and sensitivity to word order and slight morphological changes, which may not fully capture semantic equivalence. In contrast, the Meteor Score offers further improvements. Besides exact matches, Meteor Score also considers morphological changes (e.g., verb tenses, noun plurals) and synonym matches. It evaluates the similarity between the target text and the reference text more flexibly through word alignment. Nonetheless, both Meteor Score and BLEU Score primarily assess the similarity of texts at the surface level.

Although Meteor Score and BLEU Score effectively evaluate surface-level similarity, in some cases, evaluating surface similarity alone is insufficient; semantic impact must also be considered. Therefore, some works introduce the semantic score \cite{munyer2023deeptextmark, yoo2023robust, abdelnabi2021adversarial, yang2023watermarking, zhang2023remark, yang2022tracing}. A common method for evaluating semantic scores is to calculate semantic embeddings using LLMs and then compare these embeddings using cosine similarity. This process can be represented by the following formula:
\begin{equation}
\mathcal{R}_{\text{se}}(W_u, W_w) =  \frac{\mathcal{M}(W_{u}) \cdot \mathcal{M}(W_{w})}{\|\mathcal{M}(W_{u})\| \times \|\mathcal{M}(W_{w})\|}
\end{equation}
where $W_{u}$ and $W_{w}$ represent the non-watermarked text and the watermarked text, respectively. The model $\mathcal{M}$ is typically a large language model optimized for text similarity. For example, \citet{munyer2023deeptextmark} used the Universal Sentence Encoder \cite{cer2018universal}, while \citet{yoo2023robust, abdelnabi2021adversarial, yang2022tracing} used Sentence-BERT \cite{reimers2019sentence}, and \citet{yang2023watermarking} used all-MiniLM-L6-v2. 

Additionally, to determine more fine-grained relationships between the original and modified watermarked texts, some works utilize LLMs pre-trained on Natural Language Inference (NLI)  \cite{yoo2023robust, yang2022tracing} tasks to judge the relationship between two sentences. For example, \citet{yoo2023robust} used RoBERTa-Large-NLI \cite{reimers2019sentence} to more accurately understand and infer complex semantic relationships between texts (Entailment Score, ES). This Entailment score not only focuses on the overall similarity between two texts but also identifies subtle semantic differences.


\subsubsection{Single Text evaluation metrics} \label{sec:eval-text-single} Unlike comparative evaluation metrics such as BLEU Score, single text evaluation metrics focus on separately scoring the quality of the original text and the watermarked text, and then comparing these scores. Currently, the quality evaluation of watermarked text mainly uses Perplexity (PPL) or direct human evaluation.

PPL is defined as the exponentiated average negative log-likelihood of a sequence. Specifically, given a text $W = \{w_1, ..., w_{N}\}$, PPL can be computed using a LLM\(\mathcal{M}\):
\begin{equation}
    \mathcal{R}_{\text{PPL}}(W) = \exp \left( -\frac{1}{N} \sum_{i=1}^{N} \log \mathcal{M}(w_i | w_1, \ldots, w_{i-1}) \right).
    \label{eq:ppl}
\end{equation}
PPL is an important metric for evaluating text coherence and fluency. Generally, lower PPL indicates higher text quality. For more accurate evaluation, larger LLMs are typically used to calculate PPL, such as GPT-2 \cite{yang2023watermarking}, GPT-3 \cite{zhao2023provable}, OPT-2.7B \cite{DBLP:conf/icml/KirchenbauerGWK23, wang2023towards}, and LLaMA-13B \cite{liu2023semantic, liu2023private}.

Although PPL can conveniently evaluate text quality using LLMs, due to the inherent flaws of PPL (such as misrating some repetitive texts) and the accuracy of LLM outputs, many works further employ human scoring for text evaluation \cite{yoo2023robust, abdelnabi2021adversarial}. It should be noted that although human scoring is generally considered more accurate, human annotators are also prone to annotation errors. Typically, multiple human annotators are required to score the same data. However, due to high annotation costs, human scoring is challenging for large-scale evaluation.

\subsection{Output Performance Evaluation for Watermarked LLM}
\label{sec:quality-llm}

In the previous section (\cref{sec:quality-text}), we discussed how to evaluate the quality of watermarked text, which mainly pertains to watermarking for existing text (\cref{sec:existing}). For the more prevalent watermarking for existing LLM (\cref{sec:llm}) in the era of LLMs, it is usually necessary to evaluate the capabilities of the watermarked LLM. This typically involves evaluation on a series of downstream tasks, such as Text Completion \cite{DBLP:conf/icml/KirchenbauerGWK23, liu2023private, liu2023semantic, zhao2023provable, liang2024watme, liu2024adaptive, lu2024entropy, ren2023robust, yoo2023advancing, wang2023towards, wouters2023optimizing, hou2023semstamp, hou2024k, gu2024on}, Code Generation \cite{fernandez2023three, lee2023wrote, lu2024entropy}, and Machine Translation  \cite{hu2023unbiased, wu2023dipmark, liu2023semantic, liu2023private}. In this section, we will detail the various specific downstream tasks and the evaluation metrics used for these tasks.

\subsubsection{Text Completion} Since most LLMs are trained using the next word prediction paradigm, text completion is a capability that all LLMs possess. Therefore, the most common task for testing LLM capabilities is text completion. The specific approach is to provide the LLM with a text prefix as a prompt, have the LLM generate the subsequent text, and then evaluate the quality of the generated text.
Currently, the evaluation of the quality of generated text typically leverages other LLMs, including PPL \cite{DBLP:conf/icml/KirchenbauerGWK23, liu2023private, liu2023semantic, zhao2023provable, liang2024watme, liu2024adaptive, lu2024entropy, ren2023robust, yoo2023advancing, wang2023towards, wouters2023optimizing, hou2023semstamp, hou2024k, gu2024on}based on the likelihood generated by LLMs, the P-SP  \cite{kirchenbauer2023reliability, yoo2023advancing} based on text similarity, and GPT-4-score \cite{cryptoeprint:2023/1661}, which uses the more powerful GPT-4 to directly score the text.

The calculation method for the PPL metric here is the same as the eq \ref{eq:ppl} mentioned in \cref{sec:eval-text-single}. Generally, when testing PPL, larger LLMs than the current LLM are used for evaluation, with common LLMs including LLaMA-13B \cite{liu2023semantic}, LLaMA-70B \cite{liu2023private}, and GPT-3 \cite{zhao2023provable}. Due to the general applicability and simplicity of PPL, calculating the PPL metric after the LLM performs a text completion task is currently the most widely adopted evaluation method.

P-SP is more similar to a semantic score, used to evaluate the semantic similarity between two texts. \citet{yoo2023advancing} used P-SP \cite{wieting2022paraphrastic} to evaluate the semantic similarity between original human-written texts and watermarked texts generated by LLMs through the text completion task using these texts' prefixes. However, since the same text prefix can generate texts with different semantics, this evaluation has not been widely adopted.

Since the PPL and P-SP can only evaluate text quality from certain perspectives, where PPL focuses on text coherence and P-SP on semantic similarity with the original text. \citet{cryptoeprint:2023/1661} adopted GPT-4 \cite{OpenAI2023GPT4TR} to assess the quality of text generated in the text completion task. Specifically, they designed a scoring prompt for GPT-4, enabling it to output an evaluation of text quality. 


\subsubsection{Code Generation} Text completion is typically used in high-entropy scenarios and is less sensitive to changes in LLM capabilities. In contrast, low-entropy tasks like code generation are more sensitive to changes in LLM capabilities when adding watermarking, as even a small error in code can lead to failure or incorrect execution. For code generation, there are usually two evaluation methods: one is based on surface form matching, such as CodeBlue \cite{guan2024codeip}  and Edit Sim \cite{tu2023waterbench}, and the other is based on actual execution accuracy, such as Pass@k \cite{fernandez2023three, lee2023wrote, lu2024entropy}.

CodeBLEU \cite{guan2024codeip} is a metric specifically designed for evaluating the performance of code generation tasks. It extends the classic BLEU metric by incorporating the unique characteristics of code to better reflect the quality of code generation. CodeBLEU considers not only the lexical match between the generated code and the reference code but also the syntactic and semantic match. Overall, CodeBLEU is a metric based on the n-gram matching degree between the reference code and the generated watermarked code.
Edit Sim \cite{tu2023waterbench}, calculates the edit distance between two pieces of code. Both CodeBLEU and Edit Sim focus only on the surface matching degree. However, code with the same execution result can be written in different ways. Therefore, a more commonly used metric is Pass@k \cite{fernandez2023three, lee2023wrote, lu2024entropy}, which selects the top k most probable code generated by the watermarked LLM and determines the probability that one of them executes correctly.

\subsubsection{Other Tasks} In addition to testing in high-entropy and low-entropy environments (text completion and code generation), many works have evaluated the capabilities of watermarked LLMs on other typical tasks. These tasks include Question Answering \cite{he2024can, tu2023waterbench, liang2024watme, fernandez2023three}, Machine Translation  \cite{hu2023unbiased, wu2023dipmark, liu2023semantic, liu2023private}, Text Summarization \cite{hu2023unbiased, wu2023dipmark, tu2023waterbench, he2024can}, Math Reasoning \cite{fernandez2023three, liang2024watme}, Knowledge Probing \cite{tu2023waterbench}, and Instruction Following \cite{tu2023waterbench}.

The Question Answering task covers a wide range. For settings like TriviaQA \cite{joshi2017triviaqa}, which normally requires short answers, the exact match is used for evaluation. For settings like ELI5 \cite{fan2019eli5}, which require longer outputs, the Rouge is generally used. The ROUGE is similar to the BLEU, as it evaluates the similarity between the reference text and the target text (watermarked text) through n-gram similarity.
For some special scenarios, such as TruthfulQA \cite{lin2021truthfulqa}, which is used to evaluate the truthfulness and accuracy of LLMs, the GPT-Truth \cite{liang2024watme} and GPT-Info \cite{liang2024watme} metrics generated by GPT-4 can be used for evaluation.

Similarly, Machine Translation and Text Summarization tasks are often used to evaluate the capabilities of watermarked LLMs. This is typically done by adding watermarks to LLMs fine-tuned specifically for these tasks (e.g., NLLB-200 \cite{costa2022no}).
Both tasks compare the watermarked text and reference text in terms of n-gram similarity (BLEU for Machine Translation, ROUGE for Text Summarization), semantic similarity (BERT-Score), and the PPL value of the watermarked text.

Additionally, some works evaluate the capabilities of watermarked LLMs by testing their performance on other tasks. For example, the Math Reasoning \cite{fernandez2023three, liang2024watme} is used to evaluate the reasoning ability of LLMs, typically using the accuracy metric to determine if the LLM answered correctly. The Knowledge Probing \cite{tu2023waterbench}, evaluates the knowledge capability of LLMs, usually through the F1 score as the metric. Moreover, the instruction following capability \cite{tu2023waterbench} of LLMs is often evaluated using GPT-Judge to compare the original LLM and the watermarked LLM in following instructions, typically assessed by win-rate metric.



\subsection{Output Diversity Evaluation for Watermarked LLM}
\label{sec:quality-diverse}

In the previous section, we primarily discussed how to evaluate the output performance of watermarked LLMs. However, another important aspect of evaluating LLM quality is evaluating the diversity of LLM outputs. Since watermarking algorithms typically prefer certain output content, watermarked LLMs often suffer from reduced output diversity, making its evaluation a significant topic. Current work mainly focuses on evaluating the output of the Text Completion task, including metrics such as Seq-Rep-N \cite{gu2024on}, Log Diversity \cite{kirchenbauer2023reliability}, Ent-3 \cite{hou2023semstamp, hou2024k}, and Sem-Ent \cite{hou2023semstamp, hou2024k}.

\subsubsection{Seq-Rep-N}  Seq-Rep-N is typically used to calculate the repetition of n-grams in a sentence. It can be obtained by calculating the ratio of the number of unique n-grams to the total number of n-grams in a sentence. Specifically, the following formula can be used: $1 - \frac{\text{Number of unique 3-grams}}{\text{Total number of 3-grams}}$
\cite{gu2024on} analyzed using the Seq-Rep-3 metric and found that the Aar \cite{aronsonpowerpoint} has significantly higher repetition compared to KGW \cite{DBLP:conf/icml/KirchenbauerGWK23} and KTH \cite{kuditipudi2023robust}. Additionally, as the required window size increases, the repetition decreases, but this sacrifices some robustness to modifications.

\subsubsection{Log Diversity}  Furthermore, \citet{kirchenbauer2023reliability} proposed the log diversity metric. This metric calculates the negative logarithm multiplied by the proportion of unique n-grams from 1 to N, rather than just the repetition ratio of a specific n-gram. The following formula can be used for calculation: $\mathcal{R}_{d} = -\log \left(1 - \prod_{n=1}^N (1 - u_n) \right),$
where $u_n$ represents the ratio of the number of unique n-grams to the total number of n-grams in a given text sequence. Similarly, \citet{kirchenbauer2023reliability} also found that log diversity increases with the window size.

\subsubsection{Ent-3} Ent-3 measures the entropy of the frequency distribution of 3-grams in the text. A higher entropy value indicates greater lexical diversity. The following formula can be used for calculation: $H = -\sum_{i} p_i \log p_i$, where $p_i$ is the probability of the $i$th 3-gram occurring. A higher Ent-3 value indicates more diverse word choices in the generated text. \citet{hou2023semstamp} shows that their algorithm can enhance robustness without reducing diversity.

\subsubsection{Sem-Ent} Semantic Entropy (Sem-Ent) measures diversity by performing clustering analysis on the semantic representations of generated text and then calculating the entropy of the cluster distribution. It uses the LLM to generate semantic representations of the text, followed by k-means clustering analysis on all semantic representations, with a specified number of cluster centers $k$. The entropy of the cluster distribution is then calculated based on the proportion of sentences in each cluster:
$H = -\sum_{i} p_i \log p_i$
where $p_i$ is the proportion of sentences in the $i$-th cluster. A higher Sem-Ent value indicates greater semantic diversity. Sem-Ent focuses more on semantic diversity rather than just surface-form diversity.

\vspace{3pt}

Notably, for clearer understanding, in Table \ref{tab:quality_metrics} we outline evaluation metrics for quality impact mentioned in \cref{sec:quality-text}, \cref{sec:quality-llm} and \cref{sec:quality-diverse}, along with their various characteristics and the algorithms evaluated on each metric.



\subsection{Untargeted Watermark Attacks}

\label{sec:untargeted-attack}
For a text watermarking algorithm, another important evaluation aspect is its robustness against watermark attacks. Table \ref{tab:watermark_attacks} lists different types of watermark attacks. These attacks may either unintentionally modify the watermarked text (untargeted) or attempt to crack the watermark generation method to remove or forge the watermark (targeted). In this seciton, we focus on untargeted watermark attacks, leaving targeted watermark attacks for the next section.

\subsubsection{Threat Model} We first introduce the threat model for untargeted watermark attacks. We assume that the user has obtained a watermarked text. The user may or may not know that the text contains a watermark, but they do not know how the watermark was embedded. The user might modify the watermarked text, possibly at the character, word, or sentence level, or they might insert the watermarked text into a longer, non-watermarked human text.
A robust text watermarking algorithm should maintain detectability as much as possible after such modifications.

\subsubsection{Character-level Attack.} \label{sec:eval-unt-char} Merely modifying characters in the text without replacing words is a basic strategy. Random character replacement is relatively easy to detect. An alternative strategy is using visually similar Unicode characters (homoglyph attacks \cite{gabrilovich2002homograph}), which are harder for humans to notice but can be mitigated through normalization techniques \cite{DBLP:conf/icml/KirchenbauerGWK23}. Therefore, performing normalization preprocessing before passing the text to the watermark detector is crucial.

Character-level attacks affect different types of watermark algorithms differently. For format-based watermark algorithms, such as those using Unicode ID replacement to embed watermarks (e.g., EasyMark \cite{sato2023embarrassingly} using Unicode 0x0020 to replace 0x2004), homoglyph attacks can be a straightforward and effective method to remove watermarks. For other watermark algorithms, the impact mainly lies in the tokenizer; after character modification, the tokenizer may break down words into different token lists. This change in tokenization results poses a challenge to the detection effectiveness of many watermark algorithms. In summary, character-level attacks have minimal impact on text quality but are relatively easy to detect and can be removed by other methods. Therefore, this is not reliable for all scenarios.

\subsubsection{Word-level Attack to Existing Text} \label{sec:eval-unt-word-exist} Word-level attacks to existing text involve inserting, deleting, or replacing words in pre-generated watermarked text \cite{abdelnabi2021adversarial, yoo2023robust, DBLP:conf/icml/KirchenbauerGWK23,zhao2023provable, kuditipudi2023robust, yang2023watermarking}. These modifications are usually done at a fixed attack rate. Among all modification methods, synonym replacement is commonly used to minimize semantic impact, preferring replacement words that cause the least difference in sentence scoring (e.g., using BERT score \cite{devlin2018bert}).

For watermarking existing text (\cref{sec:existing}), word deletion is the most effective way. A deletion rate below 0.1 has minimal impact, but exceeding 0.3 can remove the watermark \cite{yang2023watermarking}. However, word deletion also has the greatest impact on overall semantics, potentially removing critical information. For watermarking LLMs (\cref{sec:llm}), some watermarking methods \cite{DBLP:conf/icml/KirchenbauerGWK23, kirchenbauer2023reliability, liu2023semantic} depend on previous tokens to determine the current token's watermark status. Word-level attacks alter both the current and preceding tokens, leading to the watermark's removal.

Since word-level modifications cannot alter the text order, the modification space is limited. Significant modifications are likely to disrupt sentence semantics. Although word-level attacks perform well in some scenarios, due to obvious quality issues, these methods may not accurately simulate real-world conditions.

\begin{table}[t]
\centering
\caption{The attack methods for different text watermarking algorithms. \textit{Perform Gran.} indicates the granularity of algorithm execution, \textit{Target Attack?} indicates whether the attack involves breaking the watermark algorithm details, \textit{Perform Stage} indicates whether the attack is performed before or after watermark text generation, \textit{Time Cmplx.} indicates time complexity, \textit{Quality Impact} indicates the impact on text quality, \textit{Tested Methods} lists the algorithms tested on each attack, and \textit{Robust Methods} refers to algorithms that show robustness under the corresponding attack, based on the analysis provided in the respective studies conducting the tests.}
\setlength{\tabcolsep}{10pt}
\resizebox{\textwidth}{!}{%
\begin{tabular}{>{\bfseries}
>{\centering\arraybackslash}m{3.6cm}
>{\centering\arraybackslash}m{2cm}@{} 
>{\centering\arraybackslash}m{1.5cm}@{} 
>{\centering\arraybackslash}m{2cm}@{} 
>{\centering\arraybackslash}m{1.5cm}@{} 
>{\centering\arraybackslash}m{1.5cm}@{} 
>{\centering\arraybackslash}m{3.5cm}
>{\centering\arraybackslash}m{3cm}}

\toprule
\makecell{\textbf{Attack Method}} & \makecell{\textbf{Perform} \\ \textbf{Gran.}} & \makecell{\textbf{Target} \\ \textbf{Attack?}} & \makecell{\textbf{Perform} \\ \textbf{Stage}} & \makecell{\textbf{Time} \\ \textbf{Cmplx.}} & \makecell{\textbf{Quality} \\ \textbf{Impact}} & \makecell{\textbf{Tested} \\ \textbf{Methods}}  & \makecell{\textbf{Robust} \\ \textbf{Methods}} \\
\midrule
\rowcolor{gray!25} \makecell[{{p{3.5cm}}}]{Homo. Attack \cite{gabrilovich2002homograph}} & Char & \ding{55} & Post-Gen & L. & L. & - & -\\
\makecell[{{p{3.5cm}}}]{Word Insertion} & Word & \ding{55} & Post-Gen & L. & H. & \cite{christ2024undetectable,DBLP:conf/icml/KirchenbauerGWK23,kuditipudi2023robust,aronsonpowerpoint, munyer2023deeptextmark,yoo2023robust,abdelnabi2021adversarial,zhang2023remark,lau2024waterfall,wu2023dipmark}& \cite{DBLP:conf/icml/KirchenbauerGWK23,munyer2023deeptextmark,yoo2023robust,abdelnabi2021adversarial,zhang2023remark,lau2024waterfall,wu2023dipmark,kuditipudi2023robust}\\
 \rowcolor{gray!25}\makecell[{{p{3.5cm}}}]{Word Deletion} & Word & \ding{55} & Post-Gen & L. & H. & \cite{christ2024undetectable,DBLP:conf/icml/KirchenbauerGWK23,kuditipudi2023robust,aronsonpowerpoint,munyer2023deeptextmark,yang2023watermarking,yoo2023robust,abdelnabi2021adversarial,zhang2023remark,lau2024waterfall,wu2023dipmark,zhao2023provable} & \cite{DBLP:conf/icml/KirchenbauerGWK23,munyer2023deeptextmark,yang2023watermarking,yoo2023robust,abdelnabi2021adversarial,zhang2023remark,lau2024waterfall,wu2023dipmark,zhao2023provable,kuditipudi2023robust}\\
\makecell[{{p{3.5cm}}}]{Word Replacement} & Word & \ding{55} & Post-Gen & L. & M. &\cite{christ2024undetectable,DBLP:conf/icml/KirchenbauerGWK23,kuditipudi2023robust,aronsonpowerpoint,munyer2023deeptextmark,yang2023watermarking,yoo2023robust,abdelnabi2021adversarial,zhang2023remark,lau2024waterfall,lee2023wrote,wu2023dipmark,zhao2023provable,wang2023towards,kirchenbauer2023reliability,liu2023semantic,liang2024watme,gu2024on,xu2024learning} & \cite{DBLP:conf/icml/KirchenbauerGWK23, aronsonpowerpoint,munyer2023deeptextmark,yoo2023robust,abdelnabi2021adversarial,zhang2023remark,lau2024waterfall,lee2023wrote,wu2023dipmark,zhao2023provable,kirchenbauer2023reliability,liu2023semantic,liang2024watme,kuditipudi2023robust,gu2024on,xu2024learning}\\
 \rowcolor{gray!25}
 \makecell[{{p{3.5cm}}}]{Emoji Attack \cite{DBLP:conf/icml/KirchenbauerGWK23}} & Word & \ding{55} & Pre-Gen & M. & M. & - & -\\
\makecell[{{p{3.5cm}}}]{Back-translation \cite{edunov2018understanding}} & Doc. & \ding{55} & Post-Gen & M. & L. & \cite{yang2023watermarking,lau2024waterfall,DBLP:conf/icml/KirchenbauerGWK23,hu2023unbiased,liu2023semantic,ren2023robust,lu2024entropy,kuditipudi2023robust}& \cite{yang2023watermarking,lau2024waterfall,liu2023semantic,ren2023robust,lu2024entropy}\\
\rowcolor{gray!25}
\makecell[{{p{3.5cm}}}]{CWRA \cite{he2024can}} & Doc. & \ding{55} & Post-Gen & M. & L. & \cite{he2024can, liu2023semantic, DBLP:conf/icml/KirchenbauerGWK23,hu2023unbiased} & \cite{he2024can} \\
\makecell[{{p{3.5cm}}}]{Syntax Transf. \cite{suresh2024watermarkingllmgeneratedcoderobust}} & Doc. & \ding{55} & Post-Gen & M. & M. & \cite{DBLP:conf/icml/KirchenbauerGWK23,zhao2023provable} & - \\
 \rowcolor{gray!25}
 \makecell[{{p{3.5cm}}}]{Paraphrasing} & Doc. & \ding{55} & Post-Gen & M. & L. &\cite{christ2024undetectable,DBLP:conf/icml/KirchenbauerGWK23,kuditipudi2023robust,aronsonpowerpoint,rizzo2016content,yang2023watermarking,zhang2023remark,lau2024waterfall,yoo2023advancing,zhao2023provable,wang2023towards,kirchenbauer2023reliability,liu2023semantic,liu2023private,ren2023robust,liu2024adaptive,hou2023semstamp,hou2024k,xu2024learning} & \cite{DBLP:conf/icml/KirchenbauerGWK23,zhang2023remark,lau2024waterfall,zhao2023provable,kirchenbauer2023reliability,liu2023semantic,liu2023private,ren2023robust,liu2024adaptive,hou2023semstamp,hou2024k,xu2024learning}\\
\makecell[{{p{3.5cm}}}]{Watermark \newline Collision \cite{luo2024lost}} & Doc. & \ding{55} & Post-Gen & M. & L. & \cite{liu2023semantic, DBLP:conf/icml/KirchenbauerGWK23, zhao2023provable} & -\\
\rowcolor{gray!25}
\makecell[{{p{3.5cm}}}]{Copy-Paste \cite{kirchenbauer2023reliability}} & Doc. & \ding{55} & Post-Gen & L. & L. &\cite{yoo2023advancing,wang2023towards,kirchenbauer2023reliability} &\cite{yoo2023advancing,kirchenbauer2023reliability} \\
\makecell[{{p{3.5cm}}}]{Spoofing Attack \cite{sadasivan2023aigenerated}} & \diagbox{}{}  & \ding{51} & Pre-Gen  & H. & \diagbox{}{} & \cite{DBLP:conf/icml/KirchenbauerGWK23, liu2023private, liu2023semantic,zhao2023provable,liu2024adaptive} & \cite{liu2024adaptive,liu2023private,liu2023semantic} \\
\rowcolor{gray!25}
\makecell[{{p{3.5cm}}}]{SCTS \cite{wu2024bypassing}} & \diagbox{}{}  & \ding{51} & Pre-Gen  & H. & \diagbox{}{} & \cite{DBLP:conf/icml/KirchenbauerGWK23,zhao2023provable,kirchenbauer2023reliability} & - \\
\makecell[{{p{3.5cm}}}]{MIP \cite{zhang2024large}} & \diagbox{}{}  & \ding{51} & Pre-Gen & H. & \diagbox{}{} & \cite{DBLP:conf/icml/KirchenbauerGWK23,zhao2023provable,kirchenbauer2023reliability} & -\\
\rowcolor{gray!25}
\makecell[{{p{3.5cm}}}]{Distillation \cite{gu2024on}} & \diagbox{}{}  & \ding{51} & Pre-Gen  & H. & \diagbox{}{} & \cite{DBLP:conf/icml/KirchenbauerGWK23,aronsonpowerpoint, kuditipudi2023robust} & -\\
\makecell[{{p{3.5cm}}}]{WS \cite{jovanovic2024watermark}} & \diagbox{}{}  & \ding{51} & Pre-Gen  & H. & \diagbox{}{} & \cite{DBLP:conf/icml/KirchenbauerGWK23, kirchenbauer2023reliability,zhao2023provable} & -\\
\rowcolor{gray!25}
\makecell[{{p{3.5cm}}}]{LLM Fine-tuning \cite{gu2024on}} & \diagbox{}{}  & \ding{55} & Pre-Gen  & H. & \diagbox{}{} & \cite{gu2024on} & -\\
\bottomrule
\end{tabular}
}

\label{tab:watermark_attacks}
\vspace{-3mm}
\end{table}

\subsubsection{Word-level Attack during Text Generation.} \label{sec:eval-unt-word-gen} 



Due to the inevitable impact of word-level attacks on text quality, especially with extensive modifications, recent research has begun exploring word-level attacks during text generation. These methods target watermarking algorithms for LLMs. A notable example is the emoji attack \cite{DBLP:conf/icml/KirchenbauerGWK23}, where the LLM generates emojis between each token, which are then removed post-generation.

For example, a user might request the LLM to insert "$\ddot\smile$" between each word, resulting in sentences like "There $\ddot\smile$ are $\ddot\smile$ some $\ddot\smile$ apples $\ddot\smile$ here". If "apples" is watermarked and its detection relies on the prefix (the emoji "$\ddot\smile$"), removing the emojis changes the prefix from "apples" to "some". For algorithms that rely on prefix generation for watermarking \cite{DBLP:conf/icml/KirchenbauerGWK23, liu2023private}, this attack would completely remove the watermark.

However, the effectiveness of emoji attacks depends on the LLM's ability to follow instructions. Advanced LLMs like GPT-4 \cite{OpenAI2023GPT4TR} and Claude can successfully execute emoji attacks, but less capable models might produce illogical outputs. Additionally, this attack is ineffective against watermarking methods that do not rely on previous tokens \cite{zhao2023provable, kuditipudi2023robust}.

\subsubsection{Paraphrasing Attack} \label{sec:eval-unt-para} 

Word-level attacks modify individual words to change or remove text watermarks, having limited impact. In contrast, document-level attacks involve more extensive content and structure changes, with paraphrasing being the most common method.

Paraphrasing attacks offer significant modifications but are harder to implement than word-level methods. Early techniques, like back-translation strategies \cite{yang2023watermarking}, can introduce errors and semantic drift. To enhance the quality of paraphrased text, specialized rewriting LLMs like Dipper \cite{krishna2023paraphrasing} have been created. With the rise of ChatGPT \cite{OpenAI2023GPT4TR}, many now use the \textit{gpt-3.5-Turbo} for paraphrasing, requiring just a simple prompt. Manual rewriting offers precise semantic retention and more natural expression but is costly \cite{kirchenbauer2023reliability}, especially for large texts.

Different watermarking algorithms respond variably to paraphrasing attacks. Format-based watermarks \cite{brassil1995electronic,sato2023embarrassingly,POR20121075,rizzo2016content} are particularly vulnerable, as LLMs often replace homographs with standard tokens. For LLM watermarking algorithms, token sequence dependency is crucial for robustness. Detection methods that do not depend on token order are more resilient \cite{zhao2023provable}, while sequence-dependent algorithms like KGW \cite{DBLP:conf/icml/KirchenbauerGWK23} are more robust when they rely less on previous tokens. Converting token ID dependency to semantic dependency can also improve robustness \cite{liu2023semantic, ren2023robust}. Notably, stronger watermarks and longer watermarked texts generally provide greater resistance to attacks. Notably, human writers are generally better at paraphrasing than LLMs \cite{kirchenbauer2023reliability}, though individual abilities vary significantly.

\subsubsection{Copy Paste Attack} \label{sec:eval-unt-cp} Unlike paraphrasing attacks, which aim to modify the watermarked text, copy-paste attacks \cite{kirchenbauer2023reliability} do not modify the current document but insert watermarked text into a substantial amount of human text. This type of attack weakens the watermark detector's effectiveness by reducing the proportion of watermarked text. Current work \cite{kirchenbauer2023reliability} shows when watermarked text accounts for only 10\% of the total text, the attack effect usually surpasses most paraphrasing attacks. If the proportion increases to 25\%, is is comparable to some paraphrasing attacks. Increasing text length can improve watermark detection reliability, especially in the context of copy-paste attacks.

Some watermark detection methods can identify copy-paste attacks. For example, \citet{kirchenbauer2023reliability} mentions a window test that calculates the watermark level of a specific text area instead of the entire text. This method is specifically designed to effectively detect watermarked text inserted into existing text, making it suitable for countering copy-paste attacks.

\subsubsection{Other Document-Level Attacks} 
\label{sec:other-doc}
In addition to paraphrasing and copy-paste attacks, there are other document-level untargeted attack methods like watermark collision \cite{luo2024lost}, cross-lingual watermark removal attacks (CWRA) \cite{he2024can}, and syntax transformation \cite{suresh2024watermarkingllmgeneratedcoderobust}.

Watermark collision \cite{luo2024lost} examines if rewritten text can be detected with a watermark when the paraphraser LLM itself carries another watermark. Experiments show varying sensitivities among watermarking algorithms to collisions, but high-intensity collisions can erase all initial watermarks \cite{DBLP:conf/icml/KirchenbauerGWK23, zhao2023provable, liu2023semantic}, leaving only the paraphraser's watermark. This indicates that watermark collision is a powerful attack method. Cross-lingual watermark removal attack (CWRA) \cite{he2024can} investigates whether watermarks remain when watermarked text is translated into another language. Current watermarking algorithms are not robust against such attacks. \citet{he2024can} developed the X-SIR algorithm, which shows some robustness against CWRA attacks. Syntax transformation \cite{suresh2024watermarkingllmgeneratedcoderobust} targets code by making syntactic transformations without changing functionality to erase watermarks. Current watermarks struggle to remain robust under these attacks.

\subsubsection{Fine-tuning Attacks} \label{sec:fine-tuning-attack} Fine-tuning attacks mainly target training-time watermarks (\cref{sec:train-watermark}). When text watermark features are hidden in the LLM's parameters, subsequent fine-tuning can likely remove these features. \citet{gu2024on} found that for global training-time watermarks, even minimal fine-tuning can remove the watermark features. Enhancing the robustness of training-time watermarks against subsequent fine-tuning remains a crucial research direction.

\subsection{Targeted Watermark Attacks}
\label{sec:targeted-attack}

Untargeted attacks are modifications or other forms of attacks without any knowledge of the watermark generation and detection methods. In contrast, targeted attacks occur when a malicious user attempts to crack the watermark generation method. Once the user has cracked the watermark generation method, they can easily remove existing watermarks \cite{DBLP:conf/icml/KirchenbauerGWK23} or forge new ones.

\subsubsection{Threat Model}  In targeted watermark attacks, the malicious user knows that an LLM contains a watermark. They may or may not know the specific type of watermark algorithm. Additionally, the malicious user has collected a large amount of watermarked text and may or may not have access to the watermark detector.
The goal of the malicious user is to infer the watermark generation method. A robust watermark algorithm should be difficult to crack.

\subsubsection{Targeted Watermark Attack for KGW} \label{sec:eval-t-kgw} Most current targeted watermark attacks focus on specific algorithms \cite{sadasivan2023aigenerated, zhang2024large, jovanovic2024watermark, wu2024bypassing}. These attacks understand the general approach of the watermark algorithm but lack specific details, such as the exact division of red-green word lists in the KGW algorithm \cite{DBLP:conf/icml/KirchenbauerGWK23}. The goal is to infer this division.

Spoofing attacks \cite{sadasivan2023aigenerated} statistically analyze word frequencies under a fixed prefix in watermarked text compared to normal text. High-frequency words are considered "green," while low-frequency words are "red." This method is effective for the KGW algorithm with a window size of 1 and for unigram watermarks \cite{zhao2023provable}. However, it struggles with larger window sizes.

Watermark Stealing (WS) \cite{jovanovic2024watermark} reverses watermarking rules by querying the watermark model's API. It splits watermarked and non-watermarked texts into small segments and analyzes their occurrence probabilities. Words appearing more frequently in watermarked texts are considered "green." WS achieves a spoofing attack success rate of over 80\% and can extract more complex watermarking algorithms \cite{kirchenbauer2023reliability}.

Self Color Testing-based Substitution (SCTS) \cite{wu2024bypassing} obtains color information through specific prompt generation. For instance, prompting the LLM to generate a string containing A and B, and noting which appears more frequently, determines green words. While it can identify some words, determining the entire red-green list is complex.

Mixed Integer Programming (MIP) \cite{zhang2024large} targets advanced watermarking schemes by stealing the green list and guiding optimization through systematic constraints. This method is more efficient than frequency-based methods. However, increasing the diversity and complexity of the green list can still hinder attackers from accurately identifying and replacing green markers.

\subsubsection{Watermark Distillation} \label{sec:eval-t-distill} The previously introduced algorithms can only crack the red-green word list of the KGW watermark algorithm \cite{DBLP:conf/icml/KirchenbauerGWK23} and cannot be used for other types of watermarks. \citet{gu2024on} studied the learnability of watermarks, i.e., whether an LLM can learn the watermarks by training directly on a large amount of watermarked text, thereby simulating the watermark.

They used methods of direct sampling-based learning and further distillation using logits. The experiment shows that, given sufficient training data, most current watermark algorithms can be learned, including KGW \cite{DBLP:conf/icml/KirchenbauerGWK23}, Aar \cite{aronsonpowerpoint}, and KTH \cite{kuditipudi2023robust} algorithms. However, this is limited to cases where the window size is relatively small. When the window size is sufficiently large (i.e., the watermark algorithm is sufficiently complex), these algorithms are still difficult to learn.

\vspace{3pt}

Notably, for clearer understanding, in Table \ref{tab:watermark_attacks} we outline all the watermark attacks mentioned in \cref{sec:untargeted-attack} and \cref{sec:targeted-attack}, along with their various characteristics and the algorithms evaluated on each attack.

\subsection{Benchmarks and Tools}
\label{sec:benchmark}
To facilitate the unified implementation and evaluation of text watermarks, some benchmarks and toolkits have been introduced, with WaterBench \cite{tu2023waterbench}, WaterJudge \cite{molenda2024waterjudge}, Mark My Words\citep{piet2023mark} and MarkLLM \cite{pan2024markllm} being notable examples.

WaterBench \cite{tu2023waterbench} is a comprehensive benchmark designed to evaluate the detectability of watermarks in LLMs and their impact on LLM capabilities. It sets a fixed watermark strength (e.g., 0.95) for each algorithm to ensure consistency. WaterBench includes nine tasks in five categories, covering different input and output lengths. The benchmark shows that most watermarks perform well in detection, especially in long-output tasks, but have poorer performance in short-output tasks. All watermarks reduce generation quality to some extent, particularly in open-ended tasks.

WaterJudge \cite{molenda2024waterjudge}, while focusing on similar evaluation aspects, places greater emphasis on evaluating the trade-off between watermark detectability and output quality. It uses the F1 score to measure detection performance and introduces an LLM-based evaluation approach to assess quality impact. This approach measures the average probability of an LLM preferring watermarked text over unwatermarked text in specific NLG tasks. WaterJudge compares different watermarking schemes by plotting them on a detectability-quality impact graph, providing a visual representation of this trade-off.

Mark My Words \cite{piet2023mark} evaluates watermarking schemes with a focus on watermark size and robustness under attacks. It defines watermark size as the number of tokens needed to detect the watermark at a 2\% false positive rate. It also measures robustness against eight simple attacks designed to remove the watermark while preserving semantic similarity.

In addition to these benchmarks, more comprehensive toolkits have emerged. MarkLLM \cite{pan2024markllm} is an open-source toolkit for LLM watermarking that provides a unified framework for implementing most existing LLM watermarking algorithms \cite{DBLP:conf/icml/KirchenbauerGWK23, liu2023semantic, liu2023private, aronsonpowerpoint, kuditipudi2023robust, zhao2023provable, lu2024entropy, lee2023wrote, he2024can}, ensuring ease of access through user-friendly interfaces. It also offers a comprehensive suite of evaluation tools covering detectability, quality, and robustness, as well as mechanism visualization to help the public better understand LLM watermarking technology.

\section{Application for Text Watermarking}
\label{sec:application}

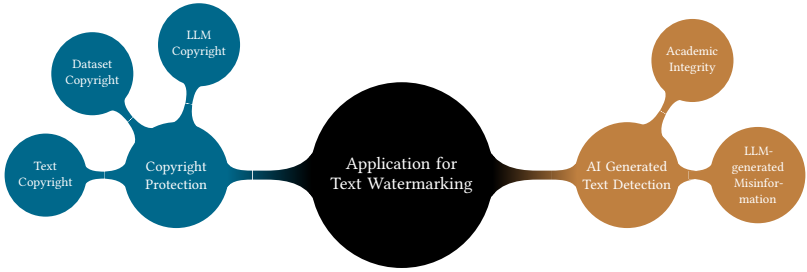
\begin{figure}[htbp]
\centering
\begin{tikzpicture}[scale=0.6, transform shape]
    \path[mindmap,concept color=black,text=white]
    node[concept] {Application for Text Watermarking}
    child[concept color=brown,grow=0]{
        node[concept] {AI Generated Text Detection} [clockwise from=60]
        child { node[concept] {Academic Integrity} [clockwise from=120]
        }
        child { node[concept] {LLM-generated Misinformation}} 
    }
    child[concept color=DeepSkyBlue4,grow=180]{ 
        node[concept] {Copyright Protection}
        child[grow=180] {node[concept] {Text Copyright}}
        child[grow=130] {node[concept] {Dataset Copyright}}
        child[grow=80] {node[concept] {LLM Copyright}}
    }
    ;
\end{tikzpicture}
\caption{Application for Text Watermarking.}
\label{fig:application}
\vspace{-3mm}
\end{figure}

In preceding sections, we outlined the implementation methods of text watermarking technologies in the era of LLMs and detailed how to thoroughly evaluate these methods. As illustrated in Figure \ref{fig:application}, this section delves into their real-world applications, focusing on two areas: copyright protection (\cref{sec:copyright}) and AI-generated text detection (\cref{sec:ai-text-detection}).

\subsection{Copyright Protection}
 \label{sec:copyright}


\subsubsection{Text Copyright}



Text copyright refers to the legal protection of original written content, ensuring creators have exclusive rights. Text watermarking technology helps safeguard copyright by detecting watermarks to identify the source of the text. The most common technology used for this purpose is format-based watermarking algorithms (\cref{sec:format}), which do not alter the text content, an important factor for many creators.

For instance, \citet{taleby2018comparative} uses layout attributes like word spacing and formatting elements such as text color and font for watermark insertion. \citet{mir2014copyright} introduced an invisible digital watermark for web content, using encrypted rules embedded in HTML. Additionally, tailored approaches are sometimes needed; \citet{iqbal2019robust} embedded watermarks in MS-Word documents using features like variables and bookmarks.

While current methods rely on format-based watermarking, the rise of LLMs suggests that integrating watermark algorithms with these models could be a promising direction for future research and applications in text copyright protection.

\subsubsection{Dataset Copyright} With the rise and widespread application of deep learning technology, the copyright of datasets has become particularly important, and protecting datasets from unauthorized use has become a critical issue. The application of text watermarking technology in this field is mainly achieved through the backdoor watermark mentioned in \ref{sec:train-watermark}.
Specifically, this method embeds specific triggers and target behaviors into the dataset. When an LLM trained on this dataset encounters the corresponding triggers, it will exhibit the target behaviors \cite{sun2023codemark, sun2022coprotector, tang2023did}.

\subsubsection{LLM Copyright} 
\label{sec:llm-copyright}
For the copyright protection of LLMs, preventing extraction attacks \cite{birch2023model, yao2024survey, patil2024can} is crucial. In these attacks, malicious users train their own LLMs using a large amount of text generated by the original LLM. Some watermarking algorithms embed watermarks in the LLM's output to prevent such attacks.

For instance, \citet{he2022protecting} proposed embedding watermarks by replacing synonyms in the generated text, choosing specific "watermark words." However, this method alters word frequency, making the watermark easier to detect and remove. To address this, \citet{he2022cater} used contextual features from parts of speech and dependency tree analysis for word replacements, keeping token frequency unchanged. \citet{zhao2023protecting} further introduced watermarks during the logits generation process (\cref{sec:logits-watermark}) by embedding periodic signals in the LLM's output logits. These signals make the watermarks more robust and covert.

Additionally, \citet{gu2024on} demonstrated that watermark algorithms like KGW \cite{DBLP:conf/icml/KirchenbauerGWK23}, Aar \cite{aronsonpowerpoint}, and KTH \cite{kuditipudi2023robust} have learnability. Moreover, data generated with these watermarks can train LLMs that carry the same watermarks. Similarly, \citet{sander2024watermarking} indicated that watermarks similar to KGW \cite{DBLP:conf/icml/KirchenbauerGWK23} give LLMs "Radioactive" properties.

However, using watermarking algorithms like KGW \cite{DBLP:conf/icml/KirchenbauerGWK23}, Aar \cite{aronsonpowerpoint}, and KTH \cite{kuditipudi2023robust} to resist model extraction attacks involves a trade-off between learnability and resistance to targeted attacks (\cref{sec:targeted-attack}). Generally, if an LLM is easy to learn from, it is also more susceptible to spoofing or stealing and can be forged more easily.

\subsection{AI Generated Text Detection}
\label{sec:ai-text-detection}

As the capabilities of LLMs grow stronger, they may be misused in an increasing number of scenarios. Typical examples include academic integrity \cite{kumar2024academic,perkins2023academic, vasilatos2023howkgpt} and LLM-generated misinformation \cite{chen2023can, 10.1145/3544548.3581318, megias2021dissimilar}. In the context of academic integrity, students might use these advanced LLMs to complete assignments, papers, or even participate in exams. In the context of LLM-generated fake news, LLMs might generate and rapidly spread false information. Current research indicates that some LLM misuses may not be totally mitigated \cite{chen2023combating, liu2024preventing}, making effective detection and tracking of LLM-generated text necessary solutions to these problems.

Theoretically, all algorithms that embed watermarks during logits generation (\cref{sec:logits-watermark}) and token sampling (\cref{sec:token-watermark}) can be applied to these scenarios. For detecting AI-generated content, some practical online services currently adopt post-generation detection methods. These methods typically use features of LLM-generated text \cite{mitchell2023detectgpt} or train classifiers to distinguish between LLM-generated text and human text \cite{lai2024adaptive}. There are also online text detection tools available, such as GPTZero \footnote{https://gptzero.me/}. This service distinguishes LLM-generated text from human text based on two features: text perplexity and burstiness. Since LLM text watermarks are added during the text generation process, there is currently no unified platform for detecting watermarked text. Building a unified watermark text detection platform should be an important future direction.

\section{Challenges and Future Directions}
\label{sec:challenges}

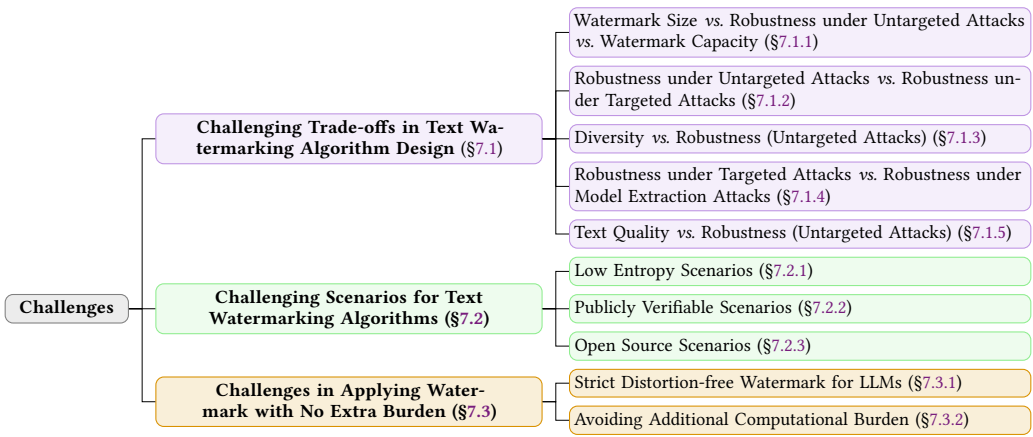
\begin{figure}[t]
\centering
\tikzset{
        my node/.style={
            draw,
            align=center,
            thin,
            text width=1.2cm, 
            rounded corners=3,
        },
        my leaf/.style={
            draw,
            align=left,
            thin,
            text width=8.5cm, 
            rounded corners=3,
        }
}
\forestset{
  every leaf node/.style={
    if n children=0{#1}{}
  },
  every tree node/.style={
    if n children=0{minimum width=1em}{#1}
  },
}
\begin{forest}
    nonleaf/.style={font=\scriptsize},
     for tree={%
        every leaf node={my leaf, font=\scriptsize},
        every tree node={my node, font=\scriptsize, l sep-=4.5pt, l-=1.pt},
        anchor=west,
        inner sep=2pt,
        l sep=10pt, 
        s sep=3pt, 
        fit=tight,
        grow'=east,
        edge={ultra thin},
        parent anchor=east,
        child anchor=west,
        if n children=0{}{nonleaf}, 
        edge path={
            \noexpand\path [draw, \forestoption{edge}] (!u.parent anchor) -- +(5pt,0) |- (.child anchor)\forestoption{edge label};
        },
        if={isodd(n_children())}{
            for children={
                if={equal(n,(n_children("!u")+1)/2)}{calign with current}{}
            }
        }{}
    }
    [\textbf{Challenges}, draw=gray, fill=gray!15, text width=1.5cm, text=black
    [\textbf{Challenging Trade-offs in Text Watermarking Algorithm Design} ({\cref{sec:challenge-trade-off}}), color=brightlavender, fill=brightlavender!15, text width=5cm, text=black
            [Watermark Size \textit{vs.} Robustness under Untargeted Attacks \textit{vs.} Watermark Capacity (\cref{sec:trade-off-1}), color=brightlavender, fill=brightlavender!15, text width=6.0cm, text=black
            ]
            [Robustness under Untargeted Attacks \textit{vs.} Robustness under Targeted Attacks (\cref{sec:trade-off-2}), color=brightlavender, fill=brightlavender!15, text width=6.0cm, text=black
            ]
            [Diversity \textit{vs.} Robustness (Untargeted Attacks) (\cref{sec:trade-off-3}), color=brightlavender, fill=brightlavender!15, text width=6.0cm, text=black
            ]
            [Robustness under Targeted Attacks \textit{vs.} Robustness under Model Extraction Attacks (\cref{sec:attack-trade-off}), color=brightlavender, fill=brightlavender!15, text width=6.0cm, text=black
            ]
            [Text Quality \textit{vs.} Robustness (Untargeted Attacks) (\cref{sec:quality-robustness-trade-off}), color=brightlavender, fill=brightlavender!15, text width=6.0cm, text=black
            ]
        ]
        [\textbf{Challenging Scenarios for Text Watermarking Algorithms (\cref{sec:challenge-senarios})}, color=lightgreen, fill=lightgreen!15, text width=5cm, text=black
            [Low Entropy Scenarios (\cref{sec:low-entropy}), color=lightgreen, fill=lightgreen!15, text width=6.0cm, text=black
            ]
            [Publicly Verifiable Scenarios (\cref{sec:public-veri}), color=lightgreen, fill=lightgreen!15, text width=6.0cm, text=black
            ]
            [Open Source Scenarios (\cref{sec:open-source}), color=lightgreen, fill=lightgreen!15, text width=6.0cm, text=black]
        ]
        [\textbf{Challenges in Applying Watermark with No Extra Burden (\cref{sec:burden})}, color=harvestgold, fill=harvestgold!15, text width=5cm, text=black
            [Strict Distortion-free Watermark for LLMs (\cref{sec:quality-challenge}), color=harvestgold, fill=harvestgold!15, text width=6.0cm, text=black
            ]
            [Avoiding Additional Computational Burden (\cref{sec:compute-challenge}), color=harvestgold, fill=harvestgold!15, text width=6.0cm, text=black
            ]
        ]
    ]
\end{forest}
\caption{Taxonomy of Challenges and Future Directions of Text Watermarking.}
\label{fig_taxonomy_of_challenges}
\vspace{-3mm}
\end{figure}

Although detailed introductions to the methods, evaluations, and application scenarios of text watermarking have been provided in previous sections, numerous challenges remain in this field. As illustrated in Figure \ref{fig_taxonomy_of_challenges}, these include challenging trade-offs in text watermarking algorithm design (\cref{sec:challenge-trade-off}), challenging scenarios for text watermarking algorithms (\cref{sec:challenge-senarios}) and challenges in applying watermark with no extra burden(\cref{sec:burden}). These challenges will be discussed in detail below.

\subsection{Challenging Trade-offs in Text Watermarking Algorithm Design}
\label{sec:challenge-trade-off}

In Section \ref{sec:evaluation}, we explore various aspects of evaluating text watermarking algorithms. However, inherent conflicts often exist between these aspects, making it extremely difficult for an algorithm to excel in all areas. We will present the existing trade-offs and their underlying reasons, and discuss potential solutions for better balancing these aspects in future work.

\subsubsection{Watermark Size, Robustness under untargeted attacks and Watermark Capacity} \label{sec:trade-off-1}
The trade-offs between watermark size (\cref{sec:eval-size}), robustness (\cref{sec:untargeted-attack}), and watermark capacity (\cref{sec:multi-bit}) are also important. Generally, improving any one of these three attributes will reduce the other two.


Specifically, as watermark capacity increases, more watermark information needs to be embedded in the text, which usually requires longer text lengths for the watermark to be detectable (less watermark size). Similarly, with higher watermark capacity, the robustness requirements of the watermarking algorithm also increase; that is, modifications to the text must not only fail to remove the watermark but also ensure that the embedded information is not altered.

The fundamental reason for contradictions among different perspectives lies in the limited suitable text space for text watermarking, usually determined by the text quality requirements. Specifically, according to Equation \color{blue}\ref{eq:quality}\color{black}, the score difference between watermarked and non-watermarked texts under the quality evaluation function $\mathcal{R}$ should be less than a threshold $\beta$. However, the number of texts meeting this criterion is limited, denoted as $|t_{\beta}|$. Since the minimal impact on text quality is a crucial feature of text watermarking algorithms, there is an upper limit for $|t_{\beta}|$ for all watermarking algorithms. Given the watermark text space of $|t_{\beta}|$, we can further analyze the conflicts between different evaluation perspectives. These trade-offs are fundamental issues in text watermarking. Although some multi-bit watermarking algorithms \cite{wang2023towards, yoo2023advancing} can mitigate this problem to some extent, these algorithms achieve multi-bit functionality at the cost of significant robustness and watermark size.

\subsubsection{Robustness under Untargeted and Targeted attacks}
\label{sec:trade-off-2}
In sections \ref{sec:untargeted-attack} and \ref{sec:targeted-attack}, we introduced how to evaluate the robustness of watermarking algorithms under Untargeted and Targeted attacks, respectively. However, there is a trade-off between these two types of robustness: methods that achieve optimal robustness in Untargeted attacks usually do not perform well in Targeted attacks, and vice versa. For most algorithms, a key factor is how many previous tokens (window size) are relied upon to generate the watermark (e.g. red-green list in KGW \cite{DBLP:conf/icml/KirchenbauerGWK23}). For watermarking algorithms based on KGW \cite{DBLP:conf/icml/KirchenbauerGWK23, hu2023unbiased, liu2023semantic, liu2023private, wu2023dipmark} and Aar \cite{aronsonpowerpoint, christ2024undetectable}, relying on more previous tokens makes the watermark generation details harder to be stolen, but it also makes the watermark easier to remove through text modification, and vice versa. 
For example, \citet{zhao2023provable} used a global red-green list split, which is considered very robust against various text modification attacks but is easily compromised by spoofing attacks. \citet{cryptoeprint:2023/1661} used a complex hash algorithm, making the watermark difficult to steal, but it has low robustness against text modifications.
This trade-off has been noted in many works \cite{liu2023semantic, gu2024on, liu2023private, jovanovic2024watermark}.

Some works attempt to mitigate this trade-off through more complex hash schemes, including self-hash \cite{kirchenbauer2023reliability}, min-hash\cite{kirchenbauer2023reliability}, and semantic hash \cite{liu2023semantic}. However, these methods only mitigate the problem to a certain extent and do not fundamentally solve it. The trade-off between robustness in Untargeted and Targeted attacks can also be referred to as the trade-off between robustness and learnability \cite{gu2024on} in some contexts \cite{gu2024on}.

\subsubsection{Diversity and Robustness under Untargeted Attacks} \label{sec:trade-off-3}  Additionally, some work \cite{kirchenbauer2023reliability, gu2024on} indicates that for LLM watermarking algorithms, there is a trade-off between output diversity and robustness against untargeted attacks. This trade-off is similar to the robustness trade-off under targeted and untargeted attacks. For some algorithms \cite{aronsonpowerpoint, DBLP:conf/icml/KirchenbauerGWK23},  more complex patterns (large window size) during design lead to more diverse rules and outputs, while simpler rule patterns limit output diversity. However, this may do not apply to all algorithms. A watermarking algorithm based on previous token hash is more susceptible to this trade-off, whereas algorithms like KTH \cite{kuditipudi2023robust} using a fixed key list experience this trade-off to a lesser extent. However, the watermark detection efficiency of KTH-type algorithms decreases, mainly because the high complexity of multi time edit distance calculation. Perhaps this trade-off should also include the time complexity of detection.
 
\subsubsection{Robustness under Targeted Attacks and Model Extraction Attacks} \label{sec:attack-trade-off} For LLM watermarking algorithms, there is a trade-off between robustness against target attacks and model extraction attacks. This was briefly mentioned in \cref{sec:llm-copyright}. The key issue is the learnability of the watermark \cite{gu2024on}. A learnable watermark can help an LLM resist model extraction attacks (by ensuring the extracted model also has the watermark features), but it also makes it easier for malicious users to extract the watermark and perform target attacks. The core of this trade-off lies in the complexity of the watermark. Complex watermark rules are more resistant to target attacks, while simple watermark rules are better at resisting model extraction attacks. This is similar to the robustness trade-off under targeted and untargeted attacks mentioned earlier.

\subsubsection{Text Quality and Robustness under untargeted Attacks} \label{sec:quality-robustness-trade-off}
Generally, enhancing the robustness of LLM text watermarking algorithms against text modifications usually means increasing the watermark strength (e.g., $\delta$ in KGW \cite{DBLP:conf/icml/KirchenbauerGWK23} or $\tau$ in Aar).  
However, this often results in larger modifications to the text, which may affect text quality. Another approach is to introduce more redundant information, making it lengthy and repetitive. Thus, there is typically a trade-off between the robustness of LLM text watermarking algorithms under untargeted attacks and text quality. However, current research suggests that this trade-off might be alleviated by sacrificing some time complexity in the watermark generation \cite{giboulot2024watermax} or detection \cite{kuditipudi2023robust}.

\subsubsection{Future Directions} Some work has been done to mitigate these trade-offs \cite{liu2023semantic, kirchenbauer2023reliability, kuditipudi2023robust}, but there is still a significant gap to achieving an optimal watermark across all aspects. Future work should better balance the above trade-offs from two perspectives: (1) Develop algorithms specifically for individual trade-offs. For example, SIR \cite{liu2023semantic} algorithm targets robustness under both untargeted and targeted attacks. (2) Design entirely new watermarking paradigms \cite{kuditipudi2023robust} that inherently achieve better balance across all trade-offs.

\subsection{Challenging Scenarios for Text Watermarking Algorithms}
\label{sec:challenge-senarios}
Current watermarking algorithms generally have good detectability and robustness. However, there are still some specific scenarios where LLM watermarking algorithms struggle to achieve excellent results. These mainly include low-entropy scenarios, publicly detectable scenarios, and open-source LLM scenarios.


\subsubsection{Low Entropy Scenarios} \label{sec:low-entropy}In low-entropy scenarios like code \cite{chen2021evaluating} or table generation \cite{li2023sequence}, embedding a highly detectable watermark is more challenging. This is mainly because these texts have strict syntactic or formatting requirements, resulting in small watermark capacity. A more in-depth explanation is that for low-entropy text, the upper limit of the watermark text space is lower, making it harder for watermark generation.

Some work has attempted to consider the impact of entropy in the watermark generation \cite{lee2023wrote} or detection \cite{lu2024entropy} process. However, they are still limited to token-level modifications. Future methods may need a stronger understanding of formatting or grammatical requirements, thereby designing semantically invariant format transformations to expand the watermark text space.



\subsubsection{Publicly Verifiable Scenarios} \label{sec:public-veri} We have already considered the Publicly Verifiable watermark scenario in \cref{sec:public-watermark}. In this scenario, the watermark detector is publicly available to users, with the goal that it remains difficult for users to forge the watermark. This poses greater challenges for the design of watermark algorithms.
Firstly, the entire watermark algorithm must have sufficient robustness against target attacks. Additionally, since the detector is public, there are more methods to perform target attacks on the algorithm, such as using the detector to reverse-engineer the generator \cite{liu2023private}. Although there has been some exploration in this area \cite{cryptoeprint:2023/1661, liu2023private}, these algorithms are typically limited by their robustness against untargeted attacks (as discussed in \cref{sec:attack-trade-off}).

Future work should explore more robust publicly verifiable watermark algorithms and investigate more potential watermark attack methods in Publicly Verifiable scenarios to further advance understanding in this field.


\subsubsection{Open Source Scenarios} \label{sec:open-source}
For LLM watermarking technology, the most challenging scenario is in the open-source scenarios. Embedding text watermarking in an open-source LLM can only be done by incorporating it into the LLM's parameters during training. This can be achieved by training on watermarked text or using some inference-time watermarking for distillation training. However, the biggest challenge in this process is the robustness to further fine-tuning. \citet{gu2024on} discovered that training-time watermarking techniques are weak in robustness to subsequent fine-tuning, and the watermark will inevitably be completely removed after sufficient fine-tuning iterations. Exploring watermarking solutions for open-source LLMs that are robust to further fine-tuning is an important direction for future research \cite{zhu2024generative}.

\subsubsection{Future Directions} The three scenarios mentioned are all highly challenging. Although some current methods attempt to adapt watermarking algorithms to these scenarios \cite{lee2023wrote, lu2024entropy, liu2023private, cryptoeprint:2023/1661, gu2024on}, they still have various limitations. Future algorithms should first explore the performance upper limits in different scenarios to design better adaptation strategies. Additionally, they should investigate other potentially more challenging scenarios, as LLMs are rapidly evolving and will likely present many new challenges in the future.


\subsection{Challenges in Applying Watermark with No Extra Burden} \label{sec:burden}

For a LLM watermarking algorithm, it is crucial and challenging to ensure that it does not introduce additional burdens. This primarily means that the LLM watermark must not cause any performance loss, nor should it add any computational burden, including in terms of time and space.


\subsubsection{Strict Distortion-free Watermark for LLMs}\label{sec:quality-challenge} Currently, many unbiased or distortion-free watermarking \cite{hu2023unbiased, aronsonpowerpoint, kuditipudi2023robust} algorithms claim not to affect LLM performance. While theoretically, these watermarking algorithms are unbiased, whether single-step unbiased decoding means no impact on capabilities is questionable. Firstly, \citet{wu2024distortion} indicates that in multi-sentence generation scenarios, these algorithms cannot be considered unbiased. Additionally, these watermarks may potentially degrade the diversity of LLM-generated text \cite{aronsonpowerpoint}. Finally, LLMs may not only use sampling-based methods (e.g. nuclear sampling) to generate text; in some scenarios, they may use beam search to generate code \cite{nijkamp2022codegen}. Whether these unbiased or distortion free watermarks \cite{kuditipudi2023robust, aronsonpowerpoint} are applicable in such scenarios is also questionable. In summary, current distortion-free watermarking works under specific sampling assumptions. More research is needed to develop general distortion-free watermarking algorithms.
\subsubsection{Avoiding Additional Computational Burden} \label{sec:compute-challenge}   While most watermarking algorithms have a low impact on LLM inference speed \cite{DBLP:conf/icml/KirchenbauerGWK23}, they only involve some hash and random number generation operations, which, if recalculated at each step, may affect latency. Some methods involve pre-calculating hash results \cite{gu2024on}, but this may have significant space overhead when the LLM's vocabulary is large or depends heavily on previous tokens. Although some algorithms theoretically have low overhead, such as Unigram \cite{zhao2023provable} and KTH \cite{kuditipudi2023robust}, they often sacrifice robustness against targeted attacks or increase the time required for watermark detection.

\subsubsection{Future Direction}In summary, to better facilitate the practical deployment of LLM watermarking algorithms, future algorithms should carefully consider their impact on LLM performance. When designing new watermarking methods, it is crucial to account for their real-world performance implications in large-scale LLM systems.

 \section{Conclusion}
 \label{conclusion}



%


%

This survey thoroughly delves into the landscape of text watermarking in the era of LLMs, encompassing its implementation, evaluation methods,  applications, challenges, and future directions.

Despite the progress made, several areas require further exploration. Future research should focus on creating advanced watermarking algorithms capable of withstanding novel attack types, especially where attackers have access to sophisticated tools and knowledge. Exploring watermarking in new applications like authenticity verification of AI-generated content in social media and journalism is crucial for maintaining the integrity and trustworthiness of digital content.

In summary, text watermarking in the era of LLMs is a rapidly evolving field. Its development will be critical in ensuring the responsible and ethical use of AI technologies.

\bibliographystyle{ACM-Reference-Format}
\bibliography{sample-base}

\appendix

\end{document}